\documentclass{article}

\usepackage{microtype}
\usepackage{graphicx}
\usepackage{caption}
\usepackage{subcaption}
\usepackage{booktabs}
\usepackage{hyperref}
\usepackage[accepted]{icml2024}
\usepackage{ushort}

\usepackage{icml2024}

\usepackage{amsmath}
\usepackage{amssymb}
\usepackage{mathtools}
\usepackage{amsthm}
\usepackage[capitalize,noabbrev]{cleveref}

\theoremstyle{plain}
\newtheorem{theorem}{Theorem}[section]

\newtheorem{lemma}[theorem]{Lemma}
\newtheorem{property}[theorem]{Property}

\theoremstyle{definition}
\newtheorem{definition}[theorem]{Definition}
\newtheorem{assumption}[theorem]{Assumption}
\theoremstyle{remark}

\DeclareMathOperator*{\argmax}{\text{arg\,max}}

\newcommand{\R}{{\mathbb R}}%

\usepackage[textsize=tiny]{todonotes}

\icmltitlerunning{Dealing with the Zero-Gradient Problem in Predict and Optimize for Convex Optimization}

\begin{document}

\twocolumn[
\icmltitle{You Shall Pass: Dealing with the Zero-Gradient Problem in Predict and Optimize for Convex Optimization}

\icmlsetsymbol{equal}{*}
\begin{icmlauthorlist}
\icmlauthor{Grigorii Veviurko}{delft}
\icmlauthor{Wendelin Boehmer}{delft}
\icmlauthor{Mathijs de Weerdt}{delft}
\end{icmlauthorlist}

\icmlaffiliation{delft}{Delft University of Technology}
\icmlcorrespondingauthor{Grigorii Veviurko}{g.veviurko@tudelft.nl}

\icmlkeywords{Predict and Optimize, Machine Learning, Convex Optimization}

\vskip 0.3in
]
\printAffiliationsAndNotice{\icmlEqualContribution} 

\begin{abstract}
Predict and optimize is an increasingly popular decision-making paradigm that employs machine learning to predict unknown parameters of optimization problems. Instead of minimizing the prediction error of the parameters, it trains predictive models using task performance as a loss function.
The key challenge to train such models is the computation of the Jacobian of the solution of the optimization problem with respect to its parameters. For linear problems, this Jacobian is known to be zero or undefined; hence, approximations are usually employed. For non-linear convex problems, however,  it is common to use the exact Jacobian. This paper demonstrates that the zero-gradient problem appears in the non-linear case as well -- the Jacobian can have a sizeable null space, thereby causing the training process to get stuck in suboptimal points.
Through formal proofs, this paper shows that smoothing the feasible set resolves this problem. Combining this insight with known techniques from the literature, such as quadratic programming approximation and projection distance regularization, a novel method to approximate the Jacobian is derived. In simulation experiments, the proposed method increases the performance in the non-linear case and at least matches the existing state-of-the-art methods for linear problems.
\end{abstract}

\section{Introduction}

\textit{Predict and optimize} (P\&O) \citep{Elmachtoub2017} is a decision-making paradigm that combines ML with mathematical programming. It considers optimization problems where some parameters are unknown and should be predicted prior to solving the problem. The P\&O approach builds upon the observation that naively training an ML algorithm to match the distribution of unknown parameters is inefficient \citep{Elmachtoub2017}, as this approach does not consider the actual task performance. Instead, P\&O aims at using the task performance as the objective function for ML models directly.

The standard way to train models in machine learning is to use gradient-based algorithms, such as stochastic gradient descent \cite{kiefer1952stochastic}. In P\&O, differentiation of the task performance involves computing the Jacobian of the solution of the optimization problem with respect to the parameters. The properties of this Jacobian depend on the problem class. For linear problems, it is known to be either a zero matrix or undefined and hence is usually approximated  \citep{Vlastelica2019, sahoo2023backpropagation, Elmachtoub2017}; in the non-linear case, the seminal work by \citet{Agrawal2019} introduced a way to compute the exact Jacobian.

In this paper, we demonstrate that the  \textit{zero-gradient} problem is not specific for the linear case, but also appears in the general non-linear convex setting. In particular, we show that the null space of the true Jacobian of a convex optimization problem depends on the number of active constraints. Consequently, the zero-gradient problem occurs during training when the solution is pulled towards the boundary of the feasible set and thereby activates the constraints. For example, it inevitably happens when the proper optimal solution lies on the boundary. Therefore, the zero-gradient problem occurs independently of the problem class and causes gradient-based methods to get stuck.

Existing methods to compute an informative approximation of the Jacobian are tailored to linear problems and cannot be applied in the general non-linear case. In this work, we propose a method that combines quadratic programming approximation 
similar to \cite{Wilder2019-ai} with projection distance regularization from \cite{Chen2021} and a novel idea of \textit{local smoothing}. The resulting algorithm has a simple geometric interpretation and is guaranteed to at least not decrease the performance.
Using multiple benchmarks, we demonstrate that this method resolves the zero-gradient problem for both linear and non-linear problems. In the former case, it performs on par with the existing methods for linear problems. In the non-linear case, it significantly outperforms the sole existing approach of using the exact Jacobian.

\section{Predict and optimize}
In this section, we first give an overview of existing research in the predict and optimize domain. Then, we define the P\&O problem and provide the necessary background.

\subsection{Related work}
The predict and optimize framework was first introduced by \citet{Elmachtoub2017}. The authors consider linear combinatorial optimization problems for which the exact Jacobian is always zero or undefined~\citep{Vlastelica2019, sahoo2023backpropagation, Wilder2019-ai}. They derive a convex sub-differentiable approximation of the task performance function to enable training.
Several other approximations are introduced in later studies: \citet{Vlastelica2019} derive a differentiable piecewise-linear approximation for the task performance, \citet{Berthet2020-rp} employ stochastic perturbations to approximate the Jacobian of combinatorial problems, and \citet{sahoo2023backpropagation} demonstrate that using simple projections on top of the predictor enables using the identity matrix as an approximation of the Jacobian.

In continuous convex optimization problems, exact differentiation of the loss function is possible. \citet{Agrawal2019} developed a differential optimization method to compute the exact Jacobian for disciplined convex programs \citep{Grant2006}. This result gave rise to new applications of P\&O in convex optimization: \citet{Uysal2021-el} applied it to the risk budgeting portfolio optimization problem, \citet{Wang2020} utilized it to learn surrogate models for P\&O problems, \citet{Donti2017} applied the method to three different real-world benchmarks. Moreover, several studies used the differential optimization technique to approximate the Jacobian of linear P\&O problems. \citet{Wilder2019-ai} construct a quadratic approximation of the problem and then use its Jacobian for backpropagation. Later, \citet{Mandi2020} improved upon this result by using logarithmic approximations. \citet{Ferber2020-aj} combined a similar idea with the cutting plane approach and used differential optimization for the combinatorial problems.

Additionally, P\&O somewhat resembles bi-level optimization (BO) \cite{sinha2017review}, as both frameworks utilize the idea of lower- and upper-level problems. Specifically, predict and optimize can be best compared with the implicit gradient methods for BO \cite{pedregosa2022, liu2021} which, however, only consider unconstrained lower-level problems. To the best of our knowledge, only one recent work \cite{Xu2023} computes implicit gradient for BO problems with lower-level constraints. Technically, their method is very similar to the one from \citet{Agrawal2019}. Because of that, the zero-gradient problem in convex differential optimization studied in our paper probably occurs in BO as well. Hence, the solution method we propose can, potentially, benefit the implicit gradient algorithms for bi-level optimization.

Summarizing, there are clear benefits of differential optimization, but whether a zero-gradient also appears in non-linear convex problems is unknown.
Moreover, if it appears, the question is what the (negative) impact is on the training process, and whether there exists a way to reduce this effect.
We aim to answer these questions with the research described in this paper.

\subsection{Problem formulation}
In this section, we introduce the P\&O problem. We refer readers to \citet{Elmachtoub2017} for further details.
In predict and optimize, we solve optimization problems of the form
\begin{equation}
   \argmax_{x} f(x, w) \text{ s. t. } x\in\mathcal{C},
   \label{eq:true-problem}
   \tag{True problem}
\end{equation}
where $x\in\mathbb{R}^n$ is the decision variable, $w\in\mathbb{R}^u$ is a vector of unknown parameters, ${f: \mathbb{R}^n \times \mathbb{R}^u \to \mathbb{R}}$ is the objective function, and $\mathcal{C}$ is the feasible set. The defining feature of this problem is that the parameters $w$ are unknown at the moment when the decision must be made. Therefore, the true optimization problem is under-defined and cannot be solved directly.

One way to deal with the unknown parameters $w$ is to use a prediction $\hat{w}$ instead. Then, the decision can be computed by solving the following problem, which we refer to as the internal problem:
\begin{equation}
   x^\ast(\hat{w}) = \argmax_{x} f(x, \hat{w}) \text{ s. t. } x\in\mathcal{C}.
   \tag{Internal problem}
\end{equation}
A commonly made assumption is that instead of $w,$ we observe a feature vector $o$ that contains some information about $w.$ Also, we have a  dataset $\mathcal{D}=\{(o_k, w_k)\},$ e.g., of historical data, which we can use to learn the relation between $w$ and $o.$ This setup enables the use of ML models for making predictions. We denote the prediction model by $\phi_\theta$, and thus we have $\hat{w} = \phi_\theta(o)$. 

The problem described above is not specific to predict and optimize. What separates the P\&O paradigm from earlier works is the approach to training the model $\phi_\theta.$ In the past, machine learning models would be trained to predict $w$ as accurately as possible, e.g., in \cite{mukhopadhyay2017prioritized}. However, the parameter prediction error is merely an artificial objective and our true goal is to derive a decision $x$ that maximizes the task performance $f(x, w).$ The main goal of the P\&O approach is to utilize this objective for training the model $\phi_\theta.$ The task performance achieved by $\phi_\theta$ on the dataset $\mathcal{D}$ can be quantified by the following loss function:
\begin{equation}
   L(\theta) = -\frac{1}{|\mathcal{D}|}\sum_{(o, w) \in \mathcal{D}} f\Big(
   x^\ast\big(\phi_\theta(o)\big), w\Big)
   \label{eq:po-loss}
\end{equation}
To train $\phi_\theta$ with a gradient-based algorithm, we need to differentiate $L$ over $\theta,$ and hence we need to compute the gradient $\nabla_{\theta}f\Big(
   x^\ast(\hat{w}), w\Big),$ where $\hat{w}=\phi_\theta(o).$ Applying the chain rule, it can be decomposed into three terms:
\begin{equation}
\begin{aligned}
  \nabla_{\theta}f\Big(x^\ast(\hat{w}), w\Big) =\!\!\nabla_{x}f\big(x^\ast(\hat{w}), w\big)\;
  \nabla_{\hat{w}} x^\ast(\hat{w})\;
  \nabla_{\theta} \hat{w}
\end{aligned}
  \label{eq:chain-rule}
\end{equation}
The second term, $\nabla_{\hat{w}}x^\ast(\hat{w}),$ is the Jacobian of the solution of the optimization problem over the prediction $\hat{w}.$ In the case of the non-linear convex optimization, it can be computed exactly \citep{Agrawal2019}.
In the next section, we show that the Jacobian $\nabla_{\hat{w}}x^\ast(\hat{w})$ can have a large null space, thereby causing the total gradient in Eq.~\ref{eq:chain-rule} to be zero even outside of the optimum.

\section{Differentiable optimization}
Without loss of generality, we consider a single instance of the problem, i.e.,~one sample $(o, w)\in\mathcal{D}.$ Everywhere in this section, we denote the prediction by $\hat{w}=\phi_\theta(o).$ Then, the decision is computed as a solution of the internal optimization problem defined as follows:
\begin{equation}
    x^\ast(\hat{w}) = \argmax_{x} f(x, \hat{w}) \text{ s.t. } x\in\mathcal{C}.
    \label{eq:int}
\end{equation}
We use $\hat{x}$ to denote the value of $x^\ast(\hat{w})$ for a given prediction $\hat{w}.$ As we are interested in convex optimization problems, we make the following assumptions:
\begin{assumption} 
The objective function $f(x, w)$ is concave and twice continuously differentiable in $x$ for any ${w}.$ \end{assumption}
\begin{assumption}
    The feasible set $\mathcal{C}$ is convex, i.e., $\{\mathcal{C}=\{x|g_i(x)\leq 0, i=1,\dots,l\},$ where $g_i(x)$ are convex differentiable functions. Moreover, for any $x\in\mathcal{C},$ the gradients $\{\nabla_{x}g_i(x)|g_i(x)=0\}$ of the active constraints are linearly independent. \footnote{As is, Assumption 2 does not allow equality constraints since they would violate linear independence property. For clarity, we use this formulation in the main body of the paper. In the appendix, we show that our results hold for the equality constraints as well.}
\end{assumption}
Additionally, we make an assumption about how $f$ depends on $w$, which holds for many real-world problems, including linear and quadratic optimization problems.

\begin{assumption}
    The objective function $f(x, w)$ is twice continuously differentiable in $w.$
\end{assumption}

Throughout this paper, we use derivatives of different objects. For clarity, we first provide an overview of them: the gradient of the true objective function over the decision, $\nabla_{x} f(\hat{x},w);$ the Jacobian of the decision over the prediction, $\nabla_{\hat{w}}x^\ast(\hat{w});$ the Jacobian of the prediction over the ML model parameters, $\nabla_{\theta}\hat{w};$ and the gradient of the predicted objective in the internal problem, $\nabla_x f(x, \hat{w}).$ In the next section, we establish some crucial properties of the Jacobian $\nabla_{\hat{w}}x^\ast(\hat{w}).$

\subsection{The zero-gradient theorem}
We begin by investigating the relation between the values of the function $x^\ast(\hat{w})$ and the gradient of the internal objective, $\nabla_{x}f(x, \hat{w}).$
Let $n_i:=\nabla_{x}g_i(\hat{x}),\,i=1,\dots,\,l$ be the normal vectors of the constraints at $\hat{x},$ Then, the KKT conditions \cite{kkt} at $\hat{x}$ state that there exist real values $\alpha_1,\ldots,\alpha_l$ such that the following holds:

\begin{equation*}
\nabla_{x}f(\hat{x}, \hat{w}) = \sum_{i=1}^{l}\alpha_in_i,\quad
\alpha_ig_i(\hat{x})=0,   
\end{equation*}

\begin{equation*}
    \alpha_i \geq 0, \quad g_i(\hat{x})\leq0,\quad i=1,\dots,l. 
\end{equation*}

Under Assumptions 1 and 2, the KKT multipliers $\alpha_i$ are uniquely defined by $\hat{w}$ and $\hat{x}.$ Thus, as $\hat{x}$ is defined by $\hat{w},$ we sometimes write $\alpha_i(\hat{w})$ to emphasize that it is, in fact, a function of $\hat{w}.$
To provide a geometrical perspective on the KKT conditions, we introduce the following definition:

\begin{definition} 
Let $x\in\mathcal{C}$ and let ${I(x)=\{i|g_i(x)=0\}}$ be the set of indices of the constraints active at $x.$ Let $n_i=\nabla_{x} g_i(x),\,\forall i\in I(x)$, be the normal vectors of these constraints. The \textnormal{gradient cone},
${G(x):=\Big\{\sum_{i\in I}\alpha_in_i|\alpha_i\geq 0 \Big\}},$ 
is the positive linear span of normal vectors $n_i.$
\label{def:zg-cone}
\end{definition}

Combining the KKT conditions with Definition \ref{def:zg-cone}, we immediately arrive at the following property:

\begin{property}
    Let $x\in\mathcal{C}$ and let $\nabla_{x}f(x, \hat{w})$ be the internal gradient at $x.$ Then, $x$ is a solution to the problem in Eq.~\ref{eq:int} if and only if \,$\forall i \in I(x),\exists \alpha_i\geq 0,$ such that 
    $\nabla_{x}f(x,\hat{w})=\sum_{i\in I(x)} \alpha_in_i\in G(x),$
    where $I(x)$ is the set of indices of active constraints, $I(x)=\{i|g_i(x)=0\}.$
    \label{property:kkt}
\end{property}

While trivial, this property provides a geometrical interpretation of the problem. Effectively, a point $x$ is a solution to the problem in Eq.~\ref{eq:int} if and only if the internal gradient at this point lies inside its gradient cone. Figure \ref{fig:zg_cone} illustrates this property. 

Before studying the Jacobian $\nabla_{\hat{w}}x^\ast(\hat{w}),$ we first need to address the question of when this Jacobian exists. Sufficient conditions for existence are given in \citet{fiacco1976sensitivity}. Under Assumptions 1-3, these conditions can be reformulated as follows:

\begin{figure}
    \centering
    \includegraphics[scale=0.15]{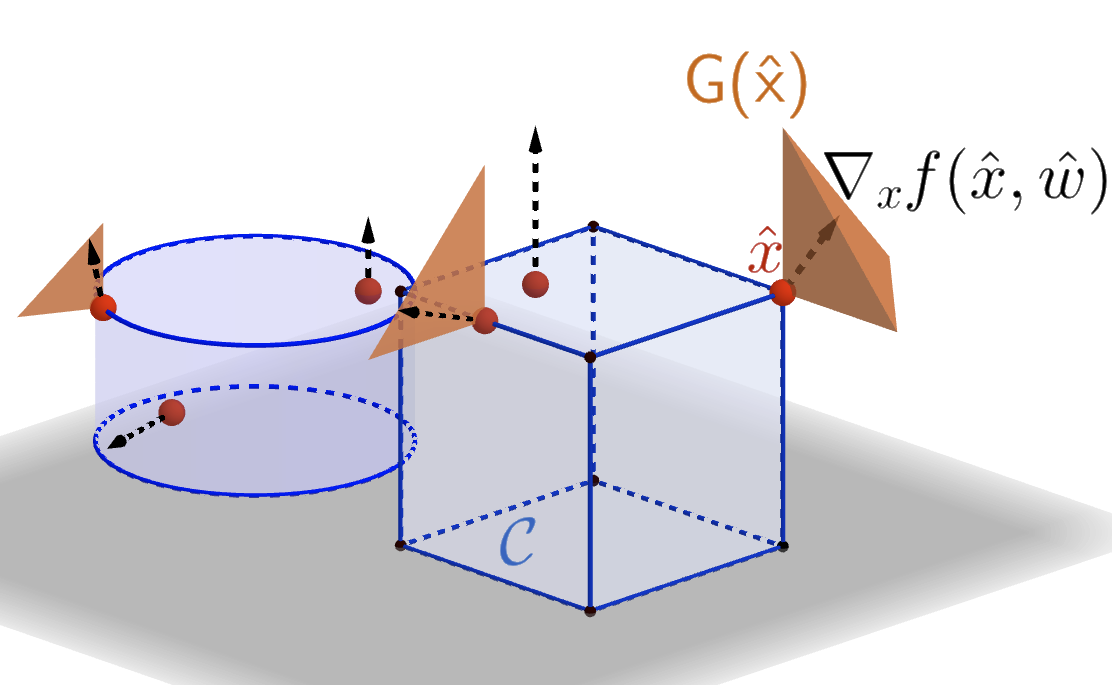}
    \caption{Gradient cones $\hat{x} + G(\hat{x})$ (orange cones) and internal gradients $\nabla_{x}f(\hat{x},\hat{w})$ (black arrows) at different points $\hat{x}$ (red dots) in different feasible sets $\mathcal{C}$ (blue cube and cylinder). The points $\hat{x}$ can not be moved in the dimensions spanned by the cones.}
    \label{fig:zg_cone}
\end{figure}
\begin{lemma}[Theorem 2.1 in \citet{fiacco1976sensitivity}]
    Let Assumptions 1-3 hold and let 
    \begin{equation*}
    \nabla_{x}f(\hat{x},\hat{w})=\sum_{i\in I(\hat{x})}\alpha_i (\hat{w})n_i
    \end{equation*}
    be the representation of the internal gradient with the normals of the active constraints. Then, suppose that the \textnormal{strict complementary slackness condition} holds, i.e., $\alpha_i(\hat{w})>0,\,\forall i\in I(\hat{x}).$
    Then, the Jacobian $\nabla_{\hat{w}}x^\ast(\hat{w})$ exists at $\hat{w}.$ Moreover, $\alpha_i(\cdot)$ is continuous around $\hat{w}$ for any $i\in I(\hat{x}).$
    \label{lemma:differentiability}
\end{lemma}
Proof of this lemma is given in \citet{fiacco1976sensitivity}. This result establishes that strict complementary slackness is sufficient for the Jacobian $\nabla_{\hat{w}}x^\ast(\hat{w})$ to exist. 
In most cases, the points that violate strict complementary slackness form a measure-zero set and hence can be neglected in practice. 

Now, we have all the necessary tools to describe the structure of the Jacobian $\nabla_{\hat{w}}x^\ast(\hat{w}).$
Suppose that the strict complementary slackness condition holds at $\hat{x}$ and hence the Jacobian exists. 
Assume that we perturb $\hat{w}$ and obtain $\hat{w}'.$ Let  $\hat{x}'=x^\ast(\hat{w}')$ denote the solution corresponding to $\hat{w}'.$ What can be said about $\hat{x}'?$ Strict complementary slackness implies that the constraints active at $\hat{x}$ will remain active at $\hat{x}'$ if the difference $\|\hat{w}' - \hat{w}\|_2^2$ is small enough. Therefore, the decision $\hat{x}'$ can only move within the tangent space of $\mathcal{C}$ at $\hat{x}$, i.e., orthogonally to all $n_i,\, i\in I(\hat{x}.)$  Hence, when more constraints are active, $\hat{x}'$ can move in less directions. Formally, we obtain the following lemma:

\begin{lemma}
Suppose that the strict complementary slackness conditions hold at $\hat{x}$ and let $\nabla_{x}f(\hat{x}, \hat{w})=\sum_{i\in I(\hat{x})}\alpha_in_i,$ $ \alpha_i> 0,\; \forall i \in I(\hat x)$
be the internal gradient. Let 
$\mathcal{N}(\hat{x})=span(\{n_i \,|\, i\in I(\hat{x})\})$
be the linear span of the gradient cone.
Then $\mathcal{N}(\hat{x})$ is contained in the left null space of $\nabla_{\hat{w}} x^\ast(\hat{w}),$ i.e., $v\,\nabla_{\hat{w}}x^\ast(\hat{w})=0,\,\forall v\in\mathcal{N}(\hat{x})$
\label{lemma:null space}
\end{lemma}
The formal proof of this result can be found in the appendix. Lemma \ref{lemma:null space} is very important, as it specifies in what directions $x^\ast(\hat{w})$ \textit{can move} as a consequence of changing $\hat{w}.$ Now, the first term in the chain rule in Eq.~\ref{eq:chain-rule}, $\nabla_{x}f(\hat{x}, w),$ specifies in what directions $x^\ast(\hat{w})$ \textit{should} move in order for the true objective to increase. Naturally, if these directions are contained in the null space of $\nabla_{\hat{w}}x^\ast(\hat{w}),$ then the total gradient in Eq.~\ref{eq:chain-rule} is zero. This observation constitutes the main theorem of this paper -- the zero-gradient theorem.

\begin{theorem}[Zero-gradient theorem] Let $\hat{w}$ be a prediction, and let $\hat{x}$ be the solution of the internal optimization problem defined in Eq.~\ref{eq:int}. Suppose that the strict complementary slackness conditions hold at $\hat{x}$ and let
$\mathcal{N}(\hat{x})=span(\{n_i \,|\, i\in I(\hat{x})\})$
be the linear span of the gradient cone at $\hat{x}.$
Then, 
$\nabla_{x} f(\hat{x}, w)\in\mathcal{N}(\hat{x})\implies
\nabla_{\theta} f(\hat{x}, w) = 0.$
\label{theorem:zg}
\end{theorem}

The proof of this theorem is obtained by simply applying Lemma \ref{lemma:null space} to the chain rule in Eq.~\ref{eq:chain-rule}. 
The theorem claims that the gradient of the P\&O loss in Eq.~\ref{eq:po-loss} can be zero in the points outside of the optimal solution. Hence, any gradient-following method ``shall not pass'' these points.
In particular, the zero-gradient phenomenon happens in such points $\hat{x}$ where the true gradient $\nabla_{x} f(\hat{x}, w)$ is contained in the space $\mathcal{N}(\hat{x})$ spanned by the gradient cone $G(\hat{x}).$ As the dimensionality of this space grows with the number of active constraints, the zero-gradient issue is particularly important for problems with a large number of constraints. 
In the worst case, $\mathcal{N}(\hat{x})$ can be as big as the whole decision space $\R^n,$ thereby making the total gradient $\nabla_{\theta}f(\hat{x}, w)$ from Eq.~\ref{eq:chain-rule} zero for any value of the true gradient $\nabla_{x}f(\hat{x}, w)$.
In the following sections, we introduce a method that resolves the zero-gradient problem and provides theoretical guarantees for its performance.

\subsection{Quadratic programming approximation}
The fundamental assumption of ``predict and optimize'' is that training $\phi_{\theta}$ using the task performance loss is better than fitting it to the true values of $w.$ Hence, the models trained with predict and optimize might output $\hat{w}$ that is significantly different from the true $w$ and yet produces good decisions. Taking this argument one step further, we claim that the objective function $f(x, \hat{w})$ in the internal optimization problem in Eq.~\ref{eq:int} does not need to be the same as the true objective $f(x, w).$ In particular, we suggest computing decisions using a simple quadratic program (QP), similar to the one proposed for linear problems by \citet{Wilder2019-ai}:
\begin{equation}
    x^\ast_{QP}(\hat{w}) = \argmax_{x} - \|x-\hat{w}\|^2_2 \text{ s.t. } x\in\mathcal{C}.
    \label{eq:qp}
\end{equation}

The reasons for this choice are manyfold. First, the internal objective $f_{QP}(x, \hat{w})=- \|x-\hat{w}\|^2_2,$ is strictly concave and hence $x^\ast_{QP}(\hat{w})$ is always uniquely-defined. Moreover, the range of $x_{QP}(\hat{w})$ is $\mathcal{C},$ i.e., $
\forall x\in\mathcal{C},\,\exists \hat{w}$ such that $x=x^\ast_{QP}(\hat{w}).$ Hence, it can represent any optimal solution. 

\citet{Wilder2019-ai} proposed to use the QP approximation to resolve the zero-gradient problem in the linear case. But, as shown in Theorem \ref{theorem:zg}, the Jacobian of $x^\ast_{QP}(w)$ can still be non-informative. 
We see the main advantage of using QP in the fact that it uses the smallest reasonable prediction vector $\hat{w}$ (one scalar per decision variable). Besides, QP approximation can represent any solution, and, as we show below, its Jacobian has a simple analytic form, which allows computing it cheaply and enables studying its theoretical properties.

The problem in Eq.~\ref{eq:qp} has a simple geometrical interpretation: the point $x=\hat{w}$ is the unconstrained maximum of $f_{QP}(x, \hat{w})$ and $x^\ast_{QP}(\hat{w})$ is its Euclidean projection on the feasible set $\mathcal{C},$ see Figure~\ref{fig:qp}. To compute the Jacobian $\nabla_{\hat{w}}\,x^\ast_{QP},$ we need to understand how perturbations of $\hat{w}$ affect $x^\ast_{QP}.$ Employing the geometrical intuition above, we obtain the following lemma:

\begin{lemma}
    Let $\hat{w}$ be a prediction and $\hat{x}$ be the optimal solution of the QP problem defined in Eq.~\ref{eq:qp}. Let the strict complementary slackness condition hold and let ${\{n_i|i\in I(\hat{x})\}}$ be the normals of the active constraints. Let
    $
        {\{e_j|j=1,\ldots,n-|I(\hat{x}|) \}}
    $
    be an orthogonal complement of vectors $\{n_i| i \in I(\hat{x})\}$ to a basis of $\R^n.$ Then, representation of the Jacobian $\nabla_{\hat{w}}x_{QP}(\hat{w})$ in the basis $\{n_i\}\cup \{e_j\}$ is a diagonal matrix. Its first $|I(\hat{x})|$ diagonal entries are zero, and the others are one.
    \label{lemma:qp-jacobian}
\end{lemma} 

Proof of this lemma can be found in the appendix. Lemma \ref{lemma:qp-jacobian} implies that the Jacobian $\nabla_{\hat{w}}x_{QP}(\hat{w})$ has a simple form and can be easily computed by hand. While providing computational benefits, this approach does not yet address the zero-gradient problem. Below, we introduce a method to compute an approximate of the Jacobian $\nabla_{\hat{w}}x_{QP}(\hat{w})$ that has at most one-dimensional null space and is guaranteed to at least not decrease the task performance.

\subsection{Local smoothing}
\begin{figure}[t]
    \begin{subfigure}[b]{0.24\textwidth}
         \includegraphics[width=4.2cm]{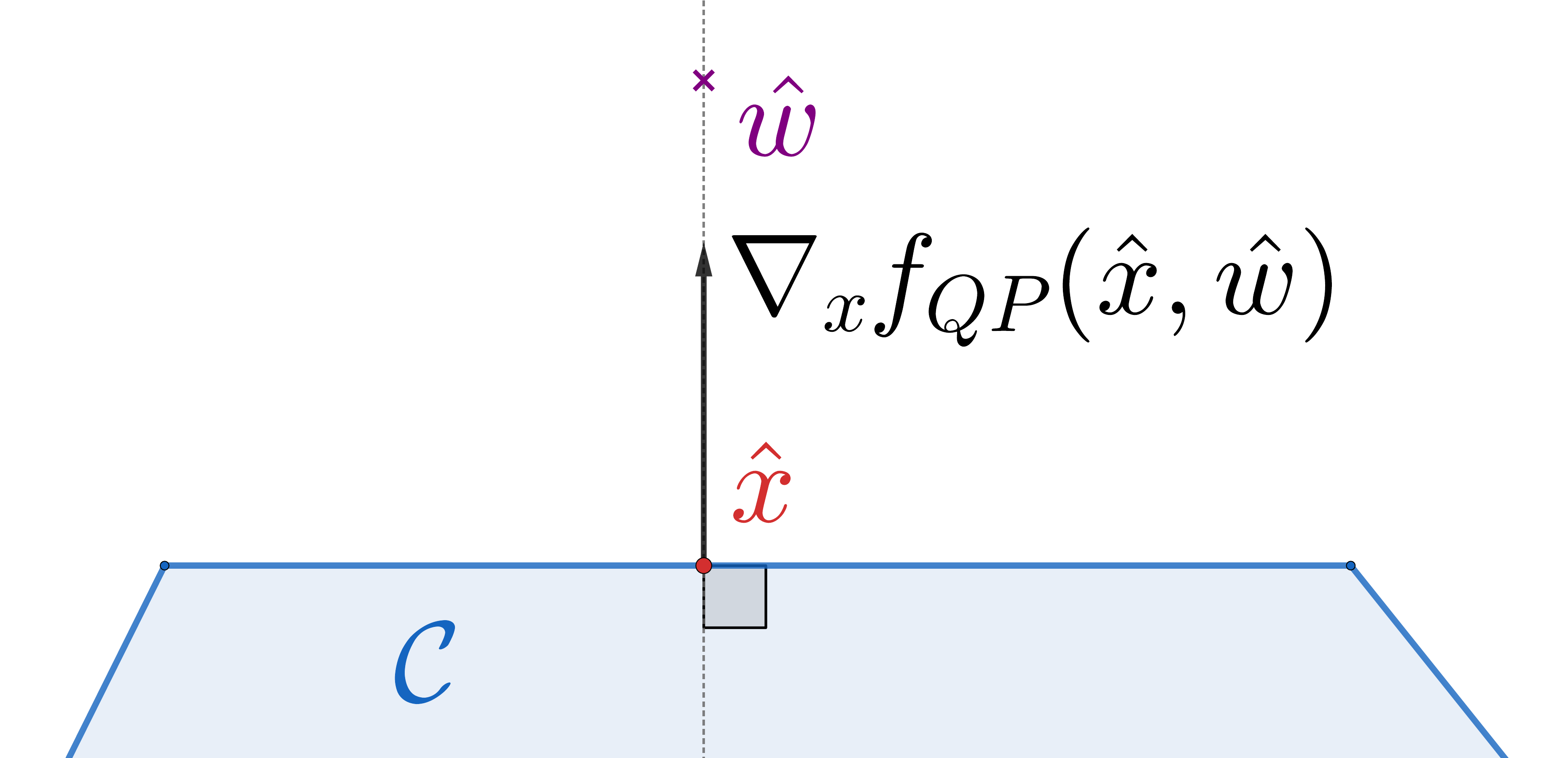}
     \end{subfigure}
    \begin{subfigure}[b]{0.22\textwidth}
         \includegraphics[width=4.2cm]{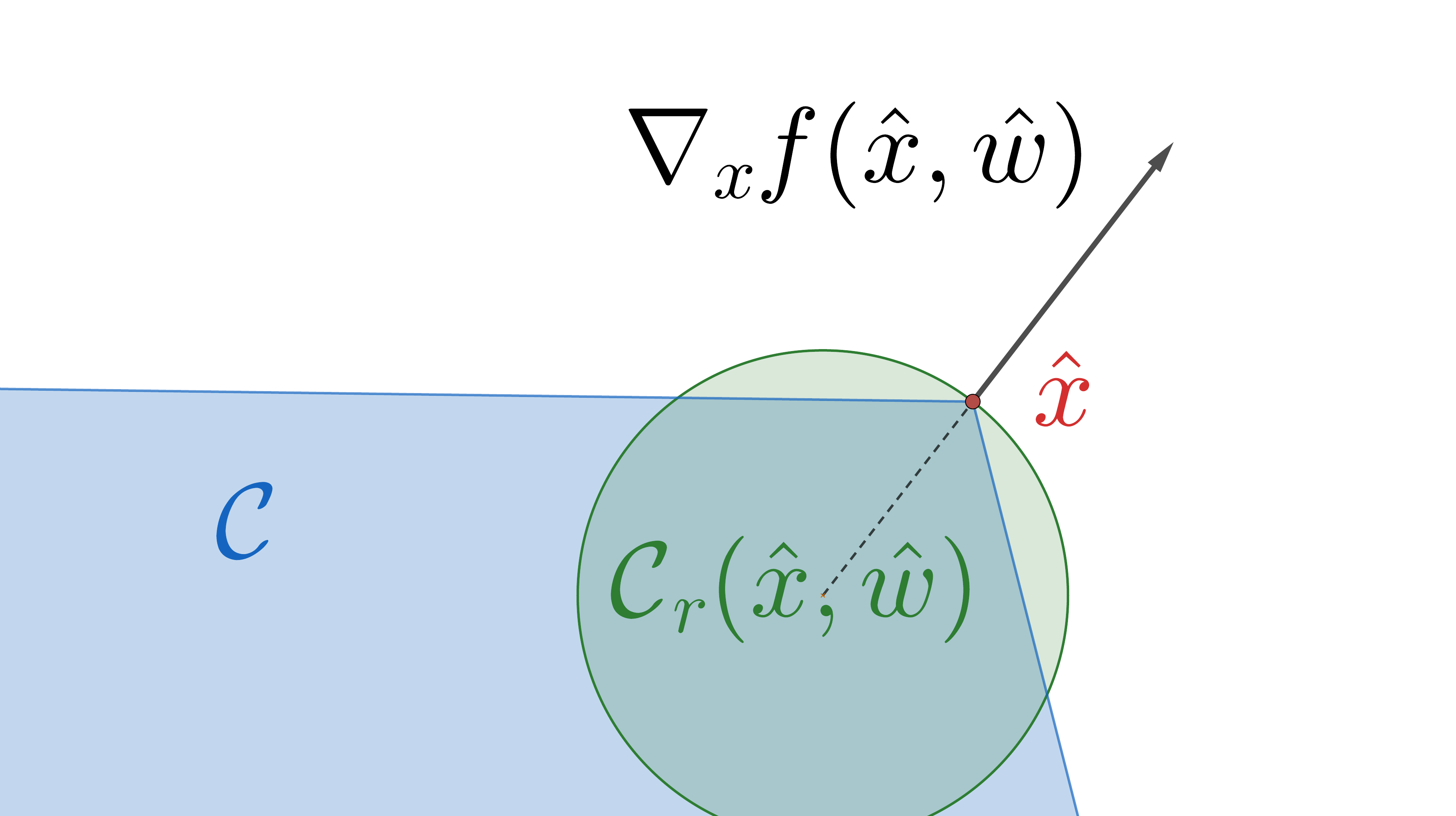}
     \end{subfigure}
    \caption{\textit{Left}: Illustration of the QP approximation. The internal gradient (black arrow) at the solution of the QP $\hat{x}$ (red point) is orthogonal to the feasible set $\mathcal{C}$ (blue area) and points towards the unconstrained maximum $\hat{w}$ (purple cross).  \textit{Right}: Illustration of the smoothed problem. The internal gradient (black arrow) is orthogonal to the smoothed feasible set $\mathcal{C}_r(\hat{x}, \hat{w})$ (green circle) at the decision $\hat{x}$ (red point).}
    \label{fig:qp}
\end{figure}
We identified that the zero-gradient problem is a fundamental issue of differential optimization that also appears in non-linear convex optimization. We showed that the size of the null space of the Jacobian $\nabla_{\hat{w}}x(\hat{w})$ depends on the number of constraints active at $\hat{x}.$ Generally, this number can be as large as the number of optimized variables $n,$ and the gradient-descent algorithms can get stuck in certain points on the boundary of the feasible set.

A potential solution to this issue is to approximate the feasible set $\mathcal{C}$ in such a way that all gradient cones become one-dimensional. 
It is known that any convex set can be approximated with a convex polytope \cite{bronstein2008approximation}, and a convex polytope can be approximated with a smooth convex set \cite{ghomi2004}. By combining these results we can approximate $\mathcal{C}$ with a smooth $\mathcal{C}'$. Then, the null space of the Jacobian of the resulting problem over $\mathcal{C}'$ becomes strictly one-dimensional. Therefore, it is possible to derive an arbitrarily close approximation of the problem such that the null space of its Jacobian is at most one-dimensional everywhere. In practice, however, this approach requires a computationally efficient way to derive an approximation of $\mathcal{C}$, and hence we leave it as a potential future work direction.
Instead, we propose a simple way to modify the feasible set -- we smooth $\mathcal{C}$ locally around the point for which we compute the Jacobian, thereby ensuring that its null space is one-dimensional. Combining this approach with the QP approximation yields a theoretically sound algorithm.

Let
$\nabla_{x}f_{QP}(\hat{x},\hat{w})=\sum_{i\in I(\hat{x})}\alpha_in_i$ be the internal gradient at $\hat{x}$ for some $\alpha_i\geq 0,\; \forall i \in I(\hat x).$ Then, we introduce the following definition:
\begin{definition}
    Let $r>0$ be a positive real number. Let $c=\hat{x} - r\frac{\nabla_{x}f_{QP}(\hat{x},\hat{w})}{\|\nabla_{x}f_{QP}(\hat{x},\hat{w})\|_2}.$
    \textnormal{The local smoothed feasible set},
    ${\mathcal{C}_r(\hat{x},\hat{w}):=\{y|y\in\R^n, \|y - c\|_2\leq r\},}$
    is a ball of radius $r$ around $c.$
    \textnormal{The local smoothed problem} $P_r(\hat{x}, \hat{w})$ with parameters $\hat{x}, \hat{w}$ is defined as 
    ${x^\ast_{r}(\hat{w}):=\argmax_{x\in\mathcal{C}_r(\hat{x}, \hat{w})}f_{QP}(x, \hat{w}).}$
\end{definition}
Figure \ref{fig:qp} shows an example of the local smoothed problem. Now, let $\hat{x}_r=x^\ast_r(\hat{w})$ denote the solution of $P_r(\hat{x}, \hat{w})$. By construction, the internal gradient at $\hat{x}_r$ lies in the one-dimensional gradient cone, and hence, by Property \ref{property:kkt}, $\hat{x}_r=\hat{x}.$
The main purpose of smoothing is to approximate the gradient in Eq.~\ref{eq:chain-rule} by substituting $\nabla_{\hat{w}}x_{QP}^\ast(\hat{w})$ with $\nabla_{\hat{w}}x^\ast_r(\hat{w}).$ We highlight that the decisions are still computed using the non-smoothed problem $x_{QP}^\ast(\hat{w})$ and $x^\ast_r(\hat{x}, \hat{w})$ is used exclusively to perform the gradient update step. In other words, we use the following expression to compute the gradient:
\begin{equation}
    \nabla_{\theta}f(x^\ast(\hat{w}), w) \approx \nabla_{x}f\big(\hat{x}, w\big)\; \nabla_{\hat{w}}x^\ast_r(\hat{w})\; \nabla_{\theta}\hat{w}
\end{equation}
Lemma \ref{lemma:qp-jacobian} prescribes the Jacobian of the smooth QP problem:
\begin{property}
Let $\hat{x}=x^\ast_{QP}(\hat{w})$ be a decision derived via QP. Suppose that the complementary slackness conditions hold for $P_r(\hat{x}, \hat{w})$ and let $e_1=\nabla_{x}f_{QP}(\hat{x}, \hat{w})$ be the internal gradient. Let $\{e_2,\ldots,e_{n}\}$ be a complement of $e_1$ to an orthogonal basis of $\R^n.$
Then, the Jacobian
$\nabla_{\hat{w}}x^\ast_r(\hat{w})$ of the locally smoothed problem expressed in the basis $\{e_1, e_2,\ldots,e_{n}\}$ is a diagonal matrix. Its first entry is zero, others are ones.
\label{property:smooth-jacobian}
\end{property}
As we see from Property \ref{property:smooth-jacobian}, the value of $r$ does not affect the Jacobian. We keep it only for notational clarity.
As $C_r(\hat{x}, \hat{w})$ is defined by a single constraint, the null space of $\nabla_{\hat{w}}x^\ast_r(\hat{x},\hat{w})$ is always one-dimensional and the zero-gradient problem only occur when the internal gradient $\nabla_{x}f_{QP}(\hat{x}, \hat{w})$ and the true gradient $\nabla_{x}f(\hat{x}, w)$ are collinear. Hence, we expect local smoothing to improve upon the zero-gradient problem. 
Property \ref{property:smooth-jacobian} reveals a connection to the work by \citet{sahoo2023backpropagation}, where the authors propose to substitute the Jacobian of linear problems by identity matrix. Then, they apply projection operators, chosen by a human, to eliminate certain directions. The local smoothing approach can be seen as an extension of this approach to the QP case. Smoothing can also be applied to other convex problems and the Jacobian will not have the simple diagonal form. This case, however, is beyond the scope of this work.

Below, we prove that the Jacobian of the smoothed QP problem yields a ``good'' direction for the gradient update.
\begin{theorem}
    Let $\hat{x}=x^\ast_{QP}(\hat{w})$ be the decision obtained via QP and let $\nabla_{\hat{w}}x^\ast_r(\hat{w})$ be the Jacobian of the local smoothed QP problem. Let $\Delta\hat{w}=\nabla_{x}f(\hat{x},w)\;\nabla_{\hat{w}}x^\ast_r(\hat{w})$ be the prediction perturbation obtained by using this Jacobian and let $\hat{w}'(t)=\hat{w} + t\Delta\hat{w}$ be the updated prediction.
    Then, for $t\to0^+,$ using $\hat{w}'(t)$ results in a non-decrease in the task performance. In other words,
    $f\big(x^\ast_{QP}(\hat{w}'(t)), w\big)\geq f\big(x^\ast_{QP}(\hat{w}), w\big).$
    \label{theorem:smooth-update}
\end{theorem}
Theorem \ref{theorem:smooth-update} shows that using local smoothing together with the QP approximation results in an analytically computable Jacobian with a one-dimensional null space that results in good gradient steps. Therefore, we are much less likely to encounter the zero-gradient problem when using this approximation. 
However, the resulting one-dimensional null space contains
the only direction that can move the prediction $\hat{w}$, and hence the decision $\hat{x},$ inside $\mathcal{C}$. This might become crucial, for example, when the optimal solution with respect to the true objective lies in the interior of $\mathcal{C}.$ To resolve this problem, we use the projection distance regularization method first suggested in \cite{Chen2021}. Specifically, we add a penalty term
\begin{equation}
    p(\hat{w}) = \alpha\|\hat{x} -\hat{w}\|_2^2,
    \label{eq:reg}
\end{equation}
where $\alpha\in\R^+$ is a hyperparameter. Minimizing this term, we push $\hat{w}$ along the null space of the Jacobian towards the feasible set and eventually move $\hat{x}$ inside $\mathcal{C}.$
\subsection{The training process}
We combine our theoretical results in the final algorithm we use to solve the P\&O problems. For each problem instance $(o, w),$ we first compute the prediction, $\hat{w}=\phi_{\theta}(o),$ and the decision using the QP approximation, $\hat{x}=x^\ast_{QP}(\hat{w}).$ Then, we obtain the achieved objective value, $f(\hat{x}, w).$ During training, we update the model parameters $\theta$ by performing the steps described in Algorithm \ref{alg:training}.

\begin{algorithm}[tb]
\caption{}
\label{alg:training}
\begin{algorithmic}
   \STATE $\hat{x}\gets x_{QP}^\ast\big(\phi(o)\big)$ \text{\quad\small\textit{Decision}}\\
   \STATE $f_x\gets \nabla_{x}f(\hat{x}, w)$\text{\quad\small \textit{True gradient}}
   \STATE $\hat{f}_x\gets\nabla_{x}f(\hat{x}, \hat{w})$\text{\quad\small \textit{Internal gradient}}
   \STATE $f^0 \gets \hat{f}_x\frac{f_x^\top \hat{f}_x}{\|\hat{f}_x\|_2}$ \text{\quad\small \textit{\;\;\;Projection on the smoothed null space}}
   \STATE $\Delta\hat{w}\gets f_x\,\nabla_{\hat{w}}x^\ast_r(\hat{w})= f_x - f^0.$ 
   \STATE $\Delta\hat{w}^{reg}\gets2\alpha(\hat{x}-\hat{w})$
   \STATE$\Delta\theta \gets (\Delta\hat{w} + \Delta\hat{w}^{reg}) \nabla_{\theta}\,\phi_\theta(o)$
   \STATE $\theta \gets \theta + \eta \Delta\theta$\text{\quad\small \textit{\quad \;Perform the gradient step}}
\end{algorithmic}
\end{algorithm}

\begin{figure*}[t!]
    \begin{subfigure}[b]{8.5cm}
    \includegraphics[width=8.5cm]{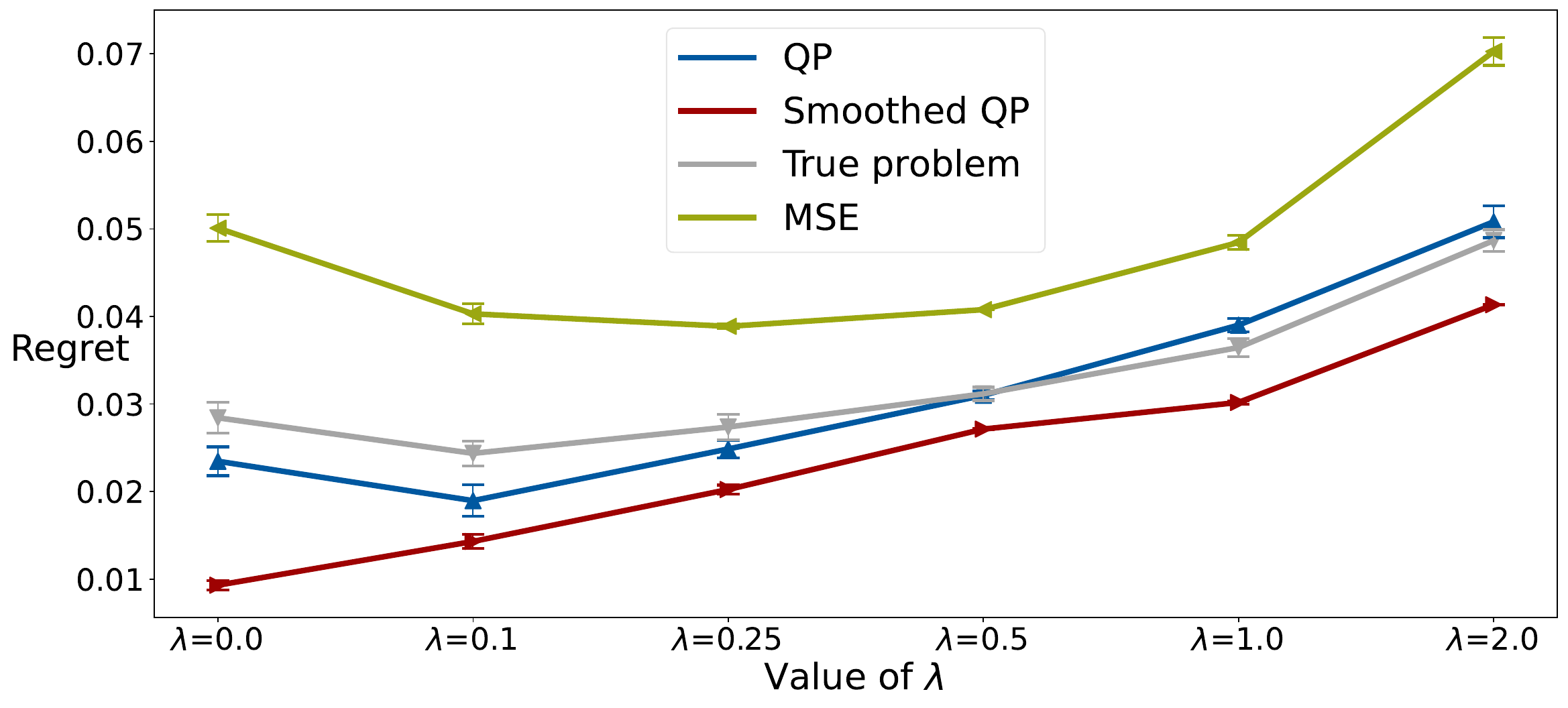}
    \caption{}
     \end{subfigure} 
     \begin{subfigure}[b]{8.5cm}
         \includegraphics[width=8.5cm]{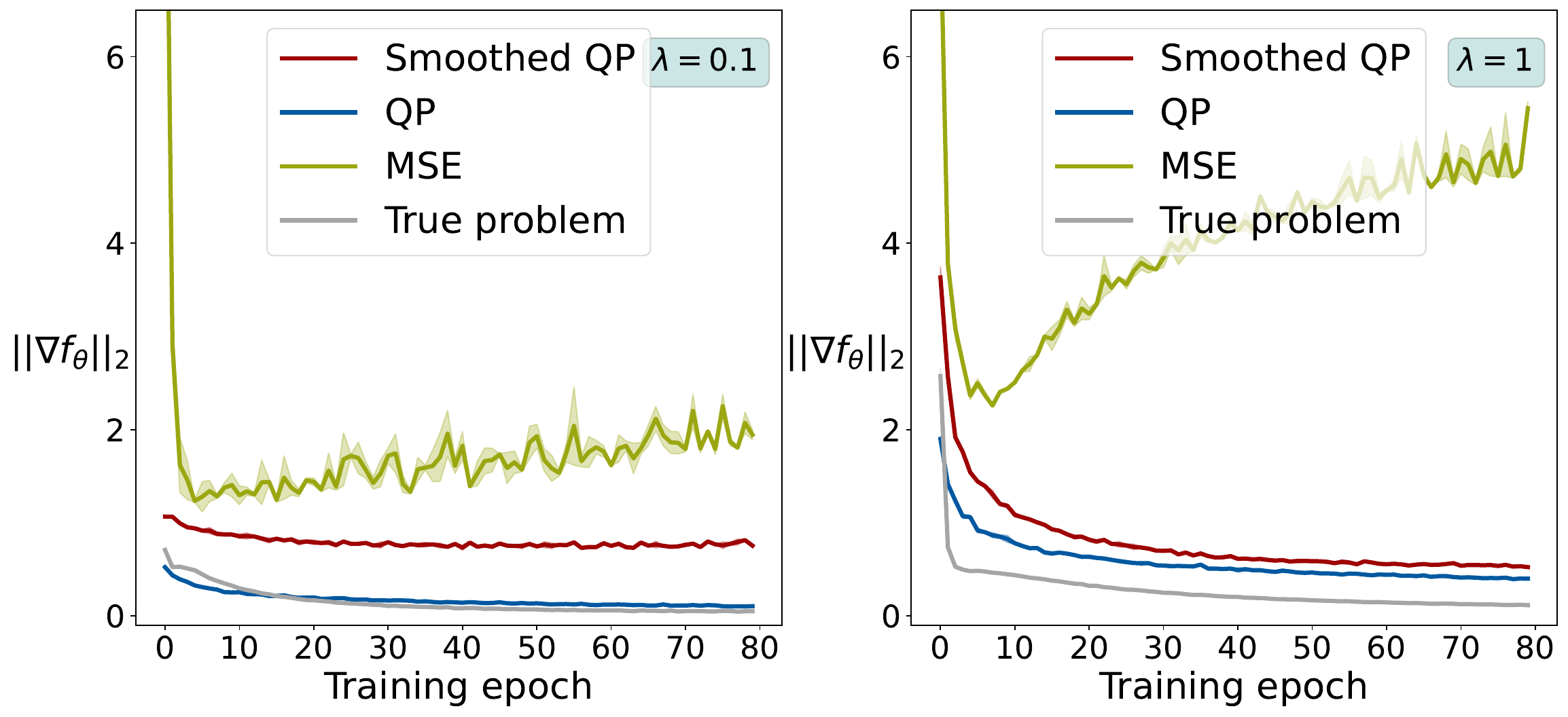}
         \caption{}
     \end{subfigure}
     \caption{Results on the standard portfolio optimization problem: (a) the final test regret for each of the algorithms for varying $\lambda$'s. The smoothed QP performs the best. \textit{(b)} Evolution of the $l_2$ norm of the gradient during training for $\lambda=0.1$ and $\lambda=1$. Unlike standalone QP and differentiation of the true problem, smoothed QP does not have the zero-gradient issue.}
     \label{fig:exp1}
\end{figure*}
\section{Experiments}
We derived the zero-gradient theorem describing when the gradient of the P\&O objective can be zero. To deal with this, we proposed a solution approach that combines QP approximation with smoothing and projection distance regularization. Below, we conduct experiments to verify our theoretical results. Specifically, we address the following: 
\begin{itemize}
    \item [-] Does the zero-gradient indeed occur in practice for non-linear P\&O problems?
    \item [-] How well does the smoothed QP method perform for linear and non-linear P\&O problems?
    \item [-] How general is QP approximation, i.e., how does it perform for different classes of the true problem?
\end{itemize}
\subsection{Portfolio optimization}
Following \citet{Wang2020}, we apply the predict and optimize framework to the Markowitz mean-variance stock market optimization problem \citet{markowitz2000mean}. In this problem, we act as an investor who seeks to maximize the immediate return but minimize the risk penalty under the budget constraint. The decision variable, $x\in\R^n,$ is a positive vector representing our investment in different securities. 
The problem is defined as follows:
\begin{equation}
    \argmax_{x} \; \underbrace{p^\top x - \lambda \, x^\top Qx}_{f(x,p,Q)}
    \quad\text{ 
    s. t.}\quad {\textstyle\sum\limits_{i=1}^n}\, x_i=1, \; x\geq0,
    \label{eq:po}
\end{equation}
where $p\in\R^n$ is the immediate return of the securities, $Q\in\R^{n\times n}$ is the positive definite matrix representing covariance between securities. The unknown parameters are defined as $w:=(p, Q).$ Importantly, $\lambda\geq0$ is a hyperparameter that represents the risk-aversion weight. We refer the readers to the appendix and to the code\footnote{Placeholder for the link to the GitHub repository} for further details regarding the implementation of this and all other benchmarks.

We consider different values of $\lambda$ from the set $\{0, 0.1, 0.25, 0.5, 1, 2\},$ in order to generate a spectrum of problems, from the linear ($\lambda=0$) to the ``strongly quadratic'' ($\lambda=2$). For smaller $\lambda$, the theory suggests that the zero-gradient problem should occur due to the true optimal solution being on the boundary of $\mathcal{C}$. This case should be resolved by local smoothing (Theorem \ref{theorem:smooth-update}). For larger values of $\lambda,$ the true optimum is often in the interior of the feasible set. In this case, the projection distance regularization (Eq.~\ref{eq:reg}) should become more important.

\begin{figure*}[t!]
    \begin{subfigure}[b]{0.24\textwidth}
         \centering
         \includegraphics[width=4.2cm]{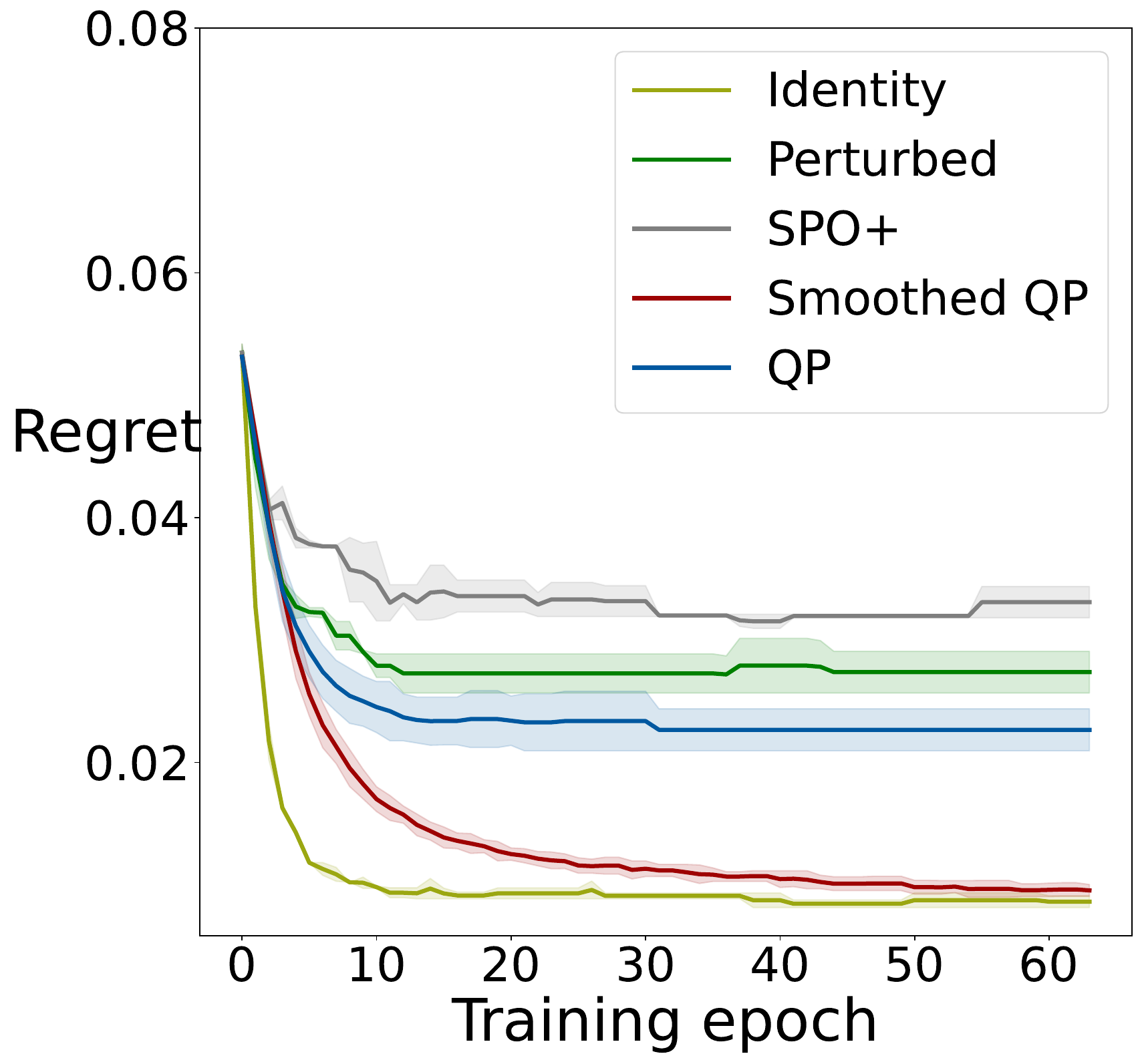}
          \caption{Linear portfolio}
     \end{subfigure} 
     \begin{subfigure}[b]{0.24\textwidth}
         \centering
         \includegraphics[width=4.2cm]{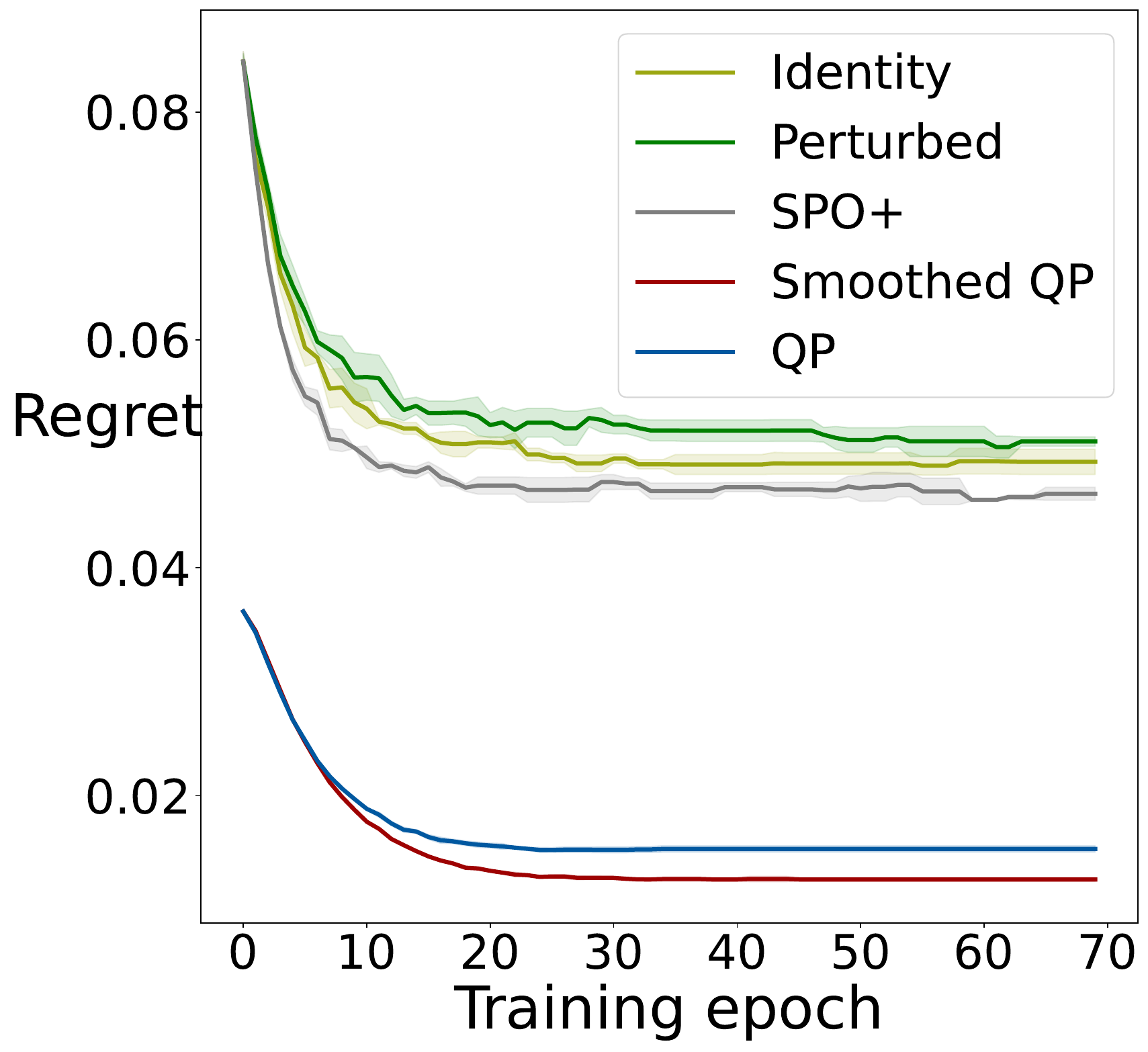}
          \caption{Portfolio with $\lambda=0.1$}
     \end{subfigure} 
     \begin{subfigure}[b]{0.24\textwidth}
         \centering
         \includegraphics[width=4.2cm]{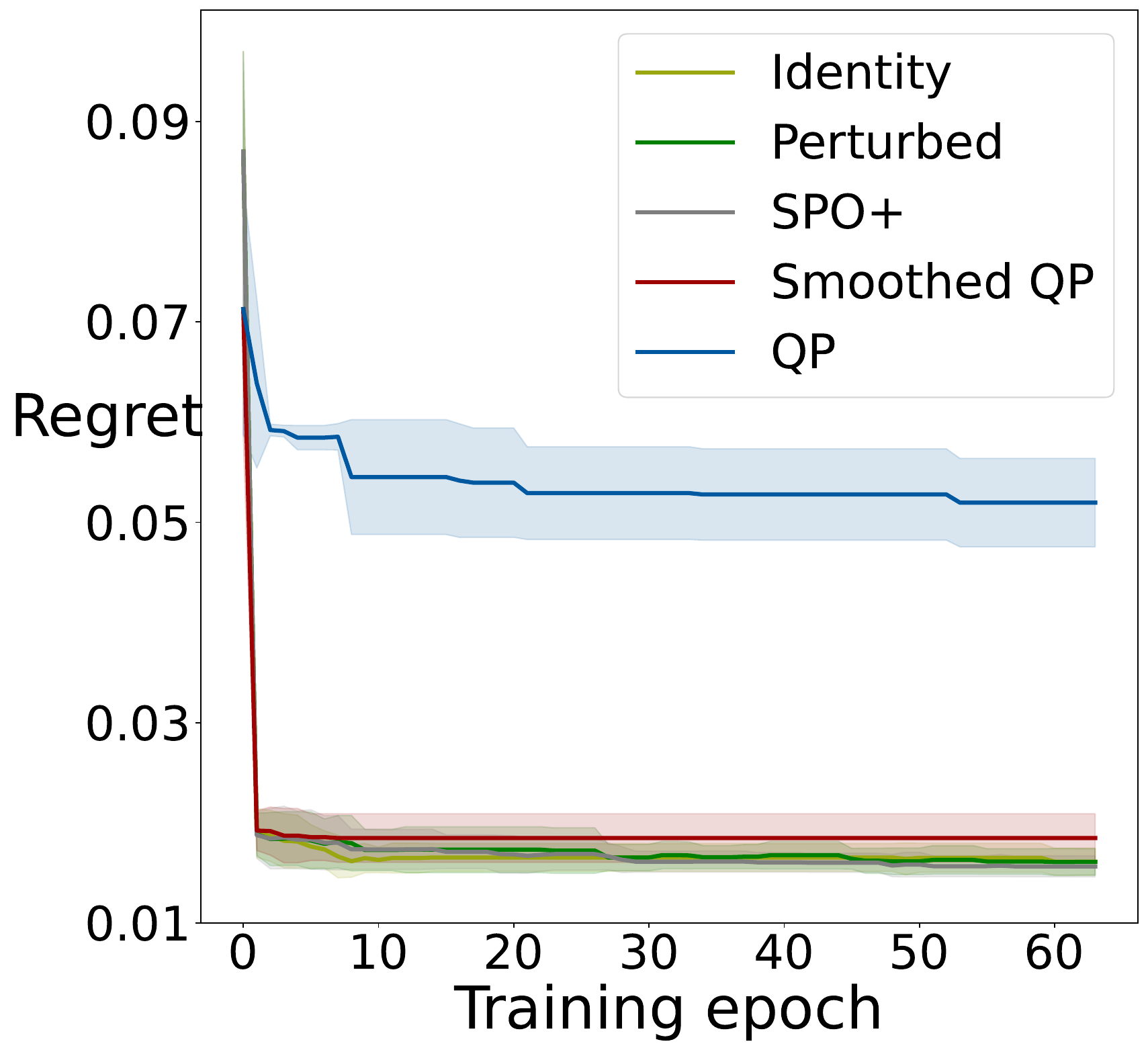}
          \caption{VOPF}
     \end{subfigure}
     \begin{subfigure}[b]{0.24\textwidth}
         \centering
         \includegraphics[width=4.2cm]{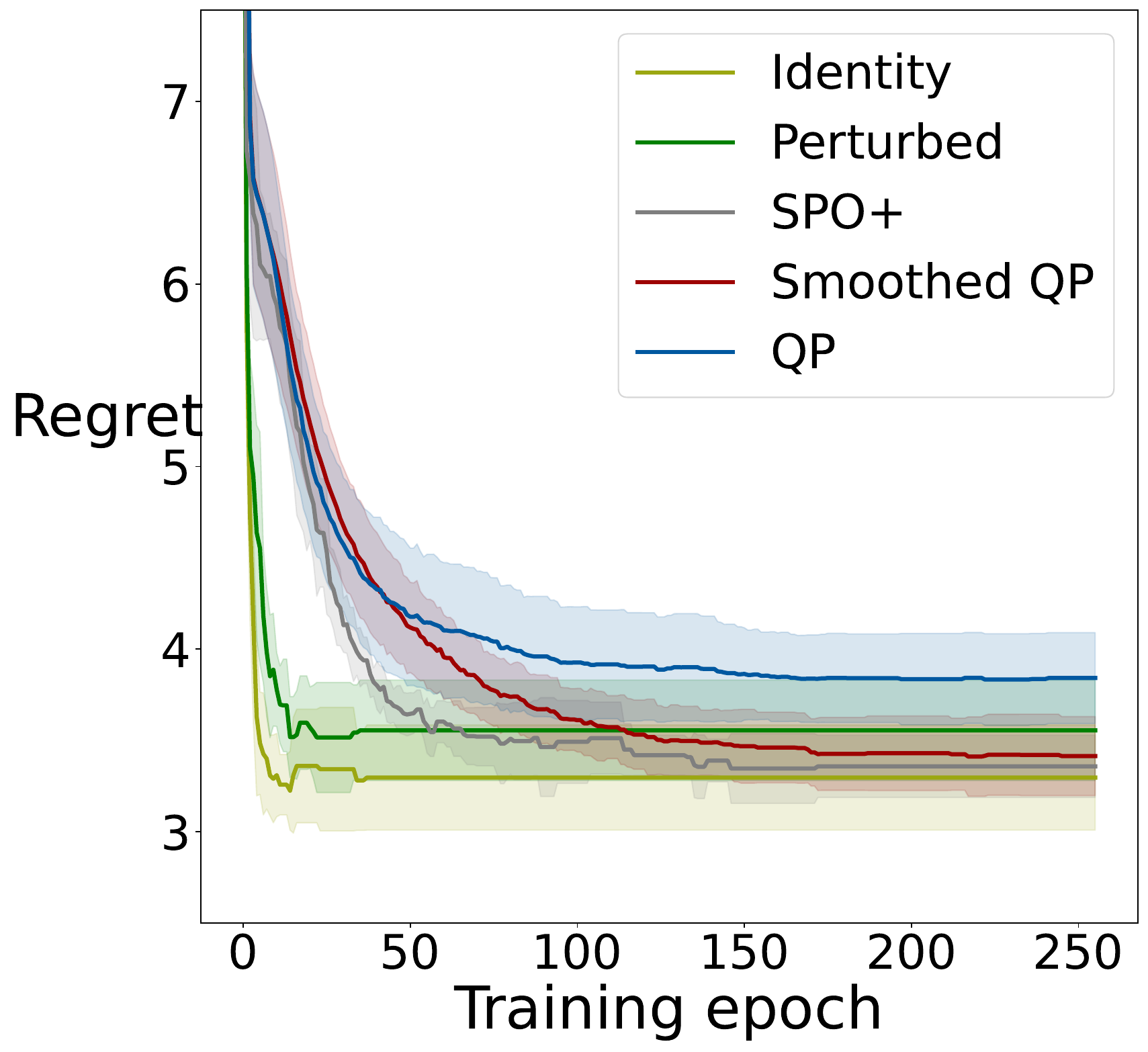}
          \caption{Knapsack}
     \end{subfigure}
     \caption{The final test regret of linear P\&O algorithms, QP approximation, and smoothed QP on four benchmark problems.}
     \label{fig:exp2}
\end{figure*}

We compare the performance of four methods: naive minimization of the mean-squared error of the prediction (labeled ``MSE''); differentiation through the true problem in Eq~\ref{eq:po} (``True problem''); differentiation through the QP approximation (``QP''); and our approach that combines QP approximation, smoothing, and projection distance regularization (``Smoothed QP''). Importantly, smoothed QP uses different values of the projection distance regularization weight $\alpha$ from Eq~\ref{eq:reg} for different $\lambda'$s. Specifically, $\alpha=0$ for $\lambda\in\{0, 0.1\},$ $\alpha=0.01$ for $\lambda\in\{0.25, 0.5\}$, and $\alpha=0.1$ for $\lambda\in\{1, 2\}$.
The remaining hyperparameter values for all methods are provided in the appendix. 
For the performance metric, we use \textit{regret} \citep{Elmachtoub2017}, defined as follows:
\begin{equation}
    \text{regret}(o, w) = f\Big(x^\ast\big(\phi_\theta(o)\big), w\Big) - \max_{x}f\big(x, w\big).
    \label{eq:regret}
\end{equation}

The results in Figure \ref{fig:exp1} demonstrate that the smoothed QP approach is dominant -- it outperforms the competitors by a significant margin across all values of $\lambda$. Figure \ref{fig:exp1} (b) suggests the reason for this result is indeed the zero-gradient problem: for the methods using the exact Jacobian (QP and true problem), the gradient norm decreases rapidly with training. In accordance with theory, this effect is more significant for smaller values of $\lambda$. In the more quadratic cases, the relative difference in performance becomes smaller. 

Figure \ref{fig:exp1} suggests that the QP approximation is sufficient in portfolio optimization. However, it might be explained by the fact that the true problem in Eq.~\ref{eq:po} is also quadratic. To gain more insight into the performance of our method in non-quadratic cases, we introduce an artificial modification of the problem's objective using the \textit{LogSumExp} function: 
\begin{equation}
    f_{lse}(x, p, Q) = -\log\sum_i e^{-p_ix_i}
    \label{eq:logsumexp}
\end{equation}
The LogSumExp function acts as a soft maximum, and the corresponding optimization problem can be interpreted as the maximization of the most profitable investment. The results in Table \ref{tab:logsumexp} demonstrate that the QP approximation outperforms the true problem in terms of regret and significantly reduces the computation time.
Moreover, smoothed QP again outperforms the other approaches, which suggests that the zero-gradient problem occurs in the LogSumExp case as well.
\begin{table}[t]
\centering
\begin{tabular}{p{2.4cm}|p{2.4cm}|p{2.4cm}}
      & {Regret} & {Runtime (sec)} \\
    \hline
    {True problem} & 0.834 $\pm$ 0.120 & 7965 $\pm$ 52\\
    {QP} &  0.506 $\pm$ 0.009 & 762 $\pm$ 52 \\
    {Smoothed QP} &  \textbf{0.438} $\pm$ \textbf{0.009} & 801 $\pm$ 54 \\
\end{tabular}
\caption{Final (normalized) test regret and training time for the different methods on the LogSumExp portfolio problem.}\label{tab:logsumexp}
\end{table}

\subsection{Comparison to linear methods}
As discussed earlier, the non-linear convex P\&O problems were considered to be solvable solely by differential optimization introduced by \citet{Agrawal2019}. Above, we showed that this is not the case as the zero-gradient phenomenon causes the training process to get stuck.
In the linear case, on the other hand, the zero-gradient is a well-known issue and there exist various methods to approximate the Jacobian of linear optimization problems.

We implemented several state-of-the-art methods for linear P\&O and compared them to the smoothed/standalone QP approaches. Specifically, we used the SPO+ loss \cite{Elmachtoub2017}, the perturbed optimizers approach \cite{Berthet2020-rp}, and the identity-with-projection method \cite{sahoo2023backpropagation}. Besides, it is worth mentioning that using the QP approximation alone is equivalent to the approach suggested by \citet{Wilder2019-ai}. We ran these methods on the linear ($\lambda=0$) and ``almost linear'' ($\lambda=0.1$) portfolio optimization problems, on the linear optimal power flow (OPF) problem, and on the continuous knapsack problem. The OPF problem is based on the linear model for the power flow in DC grids (\cite{Li2017, veviurko2022surrogate}; the knapsack is adapted from the PyEPO package \cite{tang2022pyepo}. For further details regarding these problems, we refer the readers to the appendix.

The results in Figure \ref{fig:exp2} demonstrate that our smoothed QP approach can compete with the existing algorithms in linear problems. Moreover, as soon as at least some non-linearity is introduced, Figure \ref{fig:exp2} (b), it becomes a clear winner.

In summary, the experiments provide significant evidence that the zero-gradient problem occurs in non-linear convex P\&O problems. The proposed smoothed QP method is shown to be effective in resolving the zero gradient in both linear and non-linear cases. In the linear, quadratic, and LogSumExp portfolio optimization, the QP approximation seems to be a reasonable choice for the internal problem as it performs at least as well as using the true problem. However, making claims about the universality of the QP approximation requires testing it on a bigger variety of convex non-linear, and non-quadratic P\&O problems.

\section{Conclusion}
In this research, we provide theoretical evidence for the zero-gradient problem in the general convex P\&O setting. Expanding upon existing work, we show that this problem is not specific to the linear case. In fact, it can occur for any problem class, as long as constraints are activated during training.
To resolve this issue, we introduce a method to compute an approximation of the Jacobian. It is done by smoothing the feasible set around the current solution, thereby reducing the null space's dimensionality to one. We prove that the combination of smoothing with the QP approximation results in the gradient update steps that do not decrease the task performance, but often allow to escape the zero-gradient cones. To enable movement along the remaining one-dimensional null space, we add a projection distance regularization term.

To support our theoretical findings, we conducted various experiments. Using several variants of the portfolio optimization problem, we demonstrated that the zero gradient indeed prevents the standard differential optimization approach from learning good solutions. The algorithm proposed in this work -- the smoothed QP method -- is shown to consistently outperform the existing alternative in the non-linear case. In the linear case, our approach matches the performance of the algorithms specifically designed to address the linear version of the zero-gradient problem.

In future work, we aim to investigate alternative approaches to smoothing, such as global smoothing of the feasible set. Moreover, we want to further investigate the applicability of QP approximation to more complex convex problems. Another interesting research direction is to investigate how our results can be used in BO, specifically in the problems that usually employ implicit gradients.
\newpage

\bibliographystyle{icml2024}
\bibliography{references}

\begin{thebibliography}{28}
\providecommand{\natexlab}[1]{#1}
\providecommand{\url}[1]{\texttt{#1}}
\expandafter\ifx\csname urlstyle\endcsname\relax
  \providecommand{\doi}[1]{doi: #1}\else
  \providecommand{\doi}{doi: \begingroup \urlstyle{rm}\Url}\fi

\bibitem[Agrawal et~al.(2019)Agrawal, Amos, Barratt, Boyd, Diamond, and Kolter]{Agrawal2019}
Agrawal, A., Amos, B., Barratt, S., Boyd, S., Diamond, S., and Kolter, J.~Z.
\newblock Differentiable convex optimization layers.
\newblock \emph{Adv. Neural Inf. Process. Syst.}, 32, 2019.

\bibitem[Berthet et~al.(2020)Berthet, Blondel, Teboul, Cuturi, Vert, and Bach]{Berthet2020-rp}
Berthet, Q., Blondel, M., Teboul, O., Cuturi, M., Vert, J.~P., and Bach, F.
\newblock Learning with differentiable perturbed optimizers.
\newblock \emph{Adv. Neural Inf. Process. Syst.}, 2020-Decem:\penalty0 1--24, 2020.

\bibitem[Bronstein(2008)]{bronstein2008approximation}
Bronstein, E.~M.
\newblock Approximation of convex sets by polytopes.
\newblock \emph{Journal of Mathematical Sciences}, 153\penalty0 (6):\penalty0 727--762, 2008.

\bibitem[Chen et~al.(2021)Chen, Donti, Baker, Kolter, and Berg{\'e}s]{Chen2021}
Chen, B., Donti, P.~L., Baker, K., Kolter, J.~Z., and Berg{\'e}s, M.
\newblock Enforcing policy feasibility constraints through differentiable projection for energy optimization.
\newblock \emph{e-Energy 2021 - Proceedings of the 2021 12th ACM International Conference on Future Energy Systems}, pp.\  199--210, 2021.

\bibitem[Donti et~al.(2017)Donti, Amos, and Kolter]{Donti2017}
Donti, P.~L., Amos, B., and Kolter, J.~Z.
\newblock Task-based end-to-end model learning in stochastic optimization.
\newblock \emph{Adv. Neural Inf. Process. Syst.}, 2017-Decem:\penalty0 5485--5495, 2017.

\bibitem[Elmachtoub \& Grigas(2017)Elmachtoub and Grigas]{Elmachtoub2017}
Elmachtoub, A.~N. and Grigas, P.
\newblock Smart" predict, then optimize".
\newblock \emph{arXiv preprint arXiv:1710.08005}, 2017.

\bibitem[Ferber et~al.(2020)Ferber, Wilder, Dilkina, and Tambe]{Ferber2020-aj}
Ferber, A., Wilder, B., Dilkina, B., and Tambe, M.
\newblock {MIPaaL}: Mixed integer program as a layer.
\newblock \emph{AAAI 2020 - 34th AAAI Conference on Artificial Intelligence}, pp.\  1504--1511, 2020.

\bibitem[Fiacco(1976)]{fiacco1976sensitivity}
Fiacco, A.~V.
\newblock Sensitivity analysis for nonlinear programming using penalty methods.
\newblock \emph{Mathematical programming}, 10\penalty0 (1):\penalty0 287--311, 1976.

\bibitem[Ghomi(2004)]{ghomi2004}
Ghomi, M.
\newblock Optimal smoothing for convex polytopes.
\newblock \emph{Bulletin of the London Mathematical Society}, 36\penalty0 (4):\penalty0 483--492, 2004.
\newblock \doi{https://doi.org/10.1112/S0024609303003059}.

\bibitem[Grant et~al.(2006)Grant, Boyd, and Ye]{Grant2006}
Grant, M., Boyd, S., and Ye, Y.
\newblock \emph{{Disciplined Convex Programming}}, pp.\  155--210.
\newblock Springer US, Boston, MA, 2006.
\newblock ISBN 978-0-387-30528-8.

\bibitem[Kiefer \& Wolfowitz(1952)Kiefer and Wolfowitz]{kiefer1952stochastic}
Kiefer, J. and Wolfowitz, J.
\newblock Stochastic estimation of the maximum of a regression function.
\newblock \emph{The Annals of Mathematical Statistics}, pp.\  462--466, 1952.

\bibitem[Kuhn \& Tucker(1951)Kuhn and Tucker]{kkt}
Kuhn, H.~W. and Tucker, A.~W.
\newblock Nonlinear programming.
\newblock \emph{In Berkeley Symposium on Mathematical Statistics and Probability}, 2:\penalty0 481--492, 1951.

\bibitem[Li et~al.(2018)Li, Liu, Wang, Low, and Mei]{Li2017}
Li, J., Liu, F., Wang, Z., Low, S.~H., and Mei, S.
\newblock {Optimal Power Flow in Stand-Alone DC Microgrids}.
\newblock \emph{IEEE Transactions on Power Systems}, 33\penalty0 (5):\penalty0 5496--5506, 2018.
\newblock ISSN 08858950.
\newblock \doi{10.1109/TPWRS.2018.2801280}.

\bibitem[Liu et~al.(2021)Liu, Liu, Zeng, and Zhang]{liu2021}
Liu, R., Liu, Y., Zeng, S., and Zhang, J.
\newblock Towards gradient-based bilevel optimization with non-convex followers and beyond, 2021.

\bibitem[Mandi \& Guns(2020)Mandi and Guns]{Mandi2020}
Mandi, J. and Guns, T.
\newblock Interior point solving for {LP-based} prediction+optimisation.
\newblock October 2020.

\bibitem[Markowitz \& Todd(2000)Markowitz and Todd]{markowitz2000mean}
Markowitz, H.~M. and Todd, G.~P.
\newblock \emph{Mean-variance analysis in portfolio choice and capital markets}, volume~66.
\newblock John Wiley \& Sons, 2000.

\bibitem[Mukhopadhyay \& Vorobeychik(2017)Mukhopadhyay and Vorobeychik]{mukhopadhyay2017prioritized}
Mukhopadhyay, A. and Vorobeychik, Y.
\newblock Prioritized allocation of emergency responders based on a continuous-time incident prediction model.
\newblock In \emph{International Conference on Autonomous Agents and MultiAgent Systems}, 2017.

\bibitem[Pedregosa(2022)]{pedregosa2022}
Pedregosa, F.
\newblock Hyperparameter optimization with approximate gradient, 2022.

\bibitem[{{QUANDL}}(2020)]{quandl}
{{QUANDL}}.
\newblock Quandl wiki prices, 2020., 2020.

\bibitem[Sahoo et~al.(2023)Sahoo, Paulus, Vlastelica, Musil, Kuleshov, and Martius]{sahoo2023backpropagation}
Sahoo, S.~S., Paulus, A., Vlastelica, M., Musil, V., Kuleshov, V., and Martius, G.
\newblock Backpropagation through combinatorial algorithms: Identity with projection works, 2023.

\bibitem[Sinha et~al.(2017)Sinha, Malo, and Deb]{sinha2017review}
Sinha, A., Malo, P., and Deb, K.
\newblock A review on bilevel optimization: From classical to evolutionary approaches and applications.
\newblock \emph{IEEE Transactions on Evolutionary Computation}, 22\penalty0 (2):\penalty0 276--295, 2017.

\bibitem[Tang \& Khalil(2022)Tang and Khalil]{tang2022pyepo}
Tang, B. and Khalil, E.~B.
\newblock Pyepo: A pytorch-based end-to-end predict-then-optimize library for linear and integer programming.
\newblock \emph{arXiv preprint arXiv:2206.14234}, 2022.

\bibitem[Uysal et~al.(2021)Uysal, Li, and Mulvey]{Uysal2021-el}
Uysal, A.~S., Li, X., and Mulvey, J.~M.
\newblock {End-to-End} risk budgeting portfolio optimization with neural networks.
\newblock July 2021.

\bibitem[Veviurko et~al.(2022)Veviurko, B{\"o}hmer, Mackay, and de~Weerdt]{veviurko2022surrogate}
Veviurko, G., B{\"o}hmer, W., Mackay, L., and de~Weerdt, M.
\newblock Surrogate dc microgrid models for optimization of charging electric vehicles under partial observability.
\newblock \emph{Energies}, 15\penalty0 (4):\penalty0 1389, 2022.

\bibitem[Vlastelica et~al.(2019)Vlastelica, Paulus, Musil, Martius, and Rol{\'\i}nek]{Vlastelica2019}
Vlastelica, M., Paulus, A., Musil, V., Martius, G., and Rol{\'\i}nek, M.
\newblock Differentiation of blackbox combinatorial solvers.
\newblock pp.\  1--19, 2019.

\bibitem[Wang et~al.(2020)Wang, Wilder, Perrault, and Tambe]{Wang2020}
Wang, K., Wilder, B., Perrault, A., and Tambe, M.
\newblock Automatically learning compact quality-aware surrogates for optimization problems.
\newblock June 2020.

\bibitem[Wilder et~al.(2019)Wilder, Dilkina, and Tambe]{Wilder2019-ai}
Wilder, B., Dilkina, B., and Tambe, M.
\newblock Melding the {Data-Decisions} pipeline: {Decision-Focused} learning for combinatorial optimization.
\newblock \emph{AAAI}, 33\penalty0 (01):\penalty0 1658--1665, July 2019.

\bibitem[Xu \& Zhu(2023)Xu and Zhu]{Xu2023}
Xu, S. and Zhu, M.
\newblock Efficient gradient approximation method for constrained bilevel optimization.
\newblock \emph{Proceedings of the AAAI Conference on Artificial Intelligence}, 37\penalty0 (10):\penalty0 12509–12517, June 2023.
\newblock ISSN 2159-5399.
\newblock \doi{10.1609/aaai.v37i10.26473}.
\newblock URL \url{http://dx.doi.org/10.1609/aaai.v37i10.26473}.

\end{thebibliography}

\newpage
\appendix
\onecolumn
\section{Proofs}
\begin{proof}[Proof of Lemma 4]
Let $\Delta\hat{w}$ denote an arbitrary direction and let $d=\nabla_{\hat{w}}\,x^\ast(\hat{w})\,\Delta\hat{w}$ be the corresponding directional derivative of the decision. The existence of $d$ is guaranteed by the strict complementary slackness conditions and Lemma 3.
Let $t\to0^+.$ Then, we have
\begin{equation*}
\hat{x}'(t):=x^\ast(\hat{w}+t\Delta\hat{w})=\hat{x}+td+o_x(t),
\end{equation*} where $o_x(t)$ is the ``little $o$'' notation, i.e., $\lim_{t \to 0^+} \frac{\|o_x(t)\|_2}{t}=0.$ To prove the lemma, we first want to show that $d^\top n_i =0,\;\forall i \in I(\hat{x}).$ Then, we will show that it implies the lemma's claim.

By definition, $n_i=\nabla_{x}g_i(\hat{x}).$ Then, since $g_i(\cdot)$ is differentiable and $g_i(\hat{x})=0,\,\forall i\in I(\hat{x}),$ we have the following first-order approximation for $g_i\big(\hat{x}'(t)\big):$
\begin{align*}
g_i\big(\hat{x}'(t)\big)=g_i\big(\hat{x}+td+o(t)\big)=\\
g_i(\hat{x}) + tn_i^\top d  + o_g(t) = tn_i^\top d + o_g(t).
\end{align*}
Since $\hat{x}'$ is the solution of the internal optimization problem, the inequality $g_i(\hat{x}'(t))\leq 0$ holds. Hence, the equation above implies that $n_i^\top d \leq 0.$
Now, we want to show that, in fact, $n_i^\top d = 0.$ For a proof by contradiction, suppose that $n_i^\top d < 0.$
Then, by definition of $o_g(t)$, there exists $\epsilon>0,$ such that 
\begin{equation*}
    0< t<\epsilon\implies g_i\big(\hat{x}'(t)\big)<0.
\end{equation*}
Now, we will to show that $g_i\big(\hat{x}'(t)\big)<0$ contradicts the complementary slackness condition at $\hat{x}.$ From Lemma 3, we know the KKT multiplier, $\alpha'_i(t):=\alpha_i(\hat{w}+t\Delta\hat{w}),$ is a continuous function of $t.$ On the one hand, from the KKT conditions, we know that $g_i\big(\hat{x}'(t)\big)<0\implies \alpha'_i(t)=0.$ Therefore, $\alpha'_i(t)=0$ for $t<\epsilon.$ Hence, we have
\begin{equation*}
    \lim_{t \to 0^+} \alpha'_i(t)=0.
\end{equation*} 
On the other hand, the continuity implies that
$\lim_{t \to 0^+} \alpha'_i(t)=\alpha'_i(0)=\alpha_i$ and, due to strict complementary slackness, $\alpha_i>0.$ Hence, we also have
\begin{equation*}
    \lim_{t \to 0^+} \alpha'_i(t)>0.
\end{equation*}

We arrived at a contradiction and therefore can claim that ${d^\top n_i=0}$ for all $n_i.$ 
Since ${\{n_i|i\in I(\hat{x})\}}$ is a basis of $\mathcal{N}(\hat{x}),$ this implies that for any direction $v\in\mathcal{N}(\hat{x})$ and for any $\Delta\hat{w},$ we have $v^\top\,\nabla_{\hat{w}}\,x^\ast(\hat{w})\,\Delta\hat{w}=0.$ 
In other words, vector $v^\top\,\nabla_{\hat{w}}\,x^\ast(\hat{w})$ is orthogonal to the whole space of $\hat{w}$ and hence it must be zero, $v^\top\,\nabla_{\hat{w}}\,x^\ast(\hat{w})=0,\,
\forall v\in\mathcal{N}(\hat{x}).$
Hence $\mathcal{N}(\hat{x})$ is contained in the left null space of $\nabla_{\hat{w}}\,x^\ast(\hat{w}).$
\end{proof}

\begin{proof}[Proof of Lemma 6]
First, consider the case when the unconstrained maximum $\hat{w}$ is in the interior of $\mathcal{C}.$ By definition of $x^\ast_{QP},$ it means that $\hat{x}=x^\ast_{QP}(\hat{w})$ is also in the interior of $\mathcal{C}$ and $\hat{x}=\hat{w}$. Then, $x^\ast_{QP}$ is the identity function around $\hat{w},$ and hence 
$x^\ast_{QP}(\hat{w}+\Delta\hat{w})=x(\hat{w})+\Delta\hat{w}$ for small enough $\Delta\hat{w}.$ Hence, $\nabla_{\hat{w}}x^\ast_{QP}(\hat{w})=I.$ Since no constraints are active in this case ($I(\hat{x})=\emptyset$), the lemma's claim holds.

Now, consider the case when some constraints are active, and thus $\hat{x}$ lies on the boundary of $\mathcal{C}.$ 
To get the exact form of the Jacobian $\nabla_{x}\,x_{QP}^\ast(\hat{w}),$ we will compute $\lim_{t\to0}x^\ast_{QP}(\hat{w}+t\Delta\hat{w})$ for all possible $\Delta\hat{w}.$ 
As in the QP case the predictions $\hat{w}$ lie in the same space as $\hat{x}$, we can do it first for $\Delta\hat{w}\in\mathcal{N}(\hat{x})$ and then for $\Delta\hat{w}\perp\mathcal{N}(\hat{x}).$

\paragraph{1. ${\Delta\hat{w}\in\mathcal{N}(\hat{x}).}$} For $\Delta\hat{w}\in\mathcal{N}(\hat{x}),$ we want to show that the corresponding directional derivative is zero. We begin by computing the internal gradient $\nabla_{x}f_{QP}(\hat{x}, \hat{w}):$ 
\begin{equation*}
\nabla_{x}f_{QP}(\hat{x}, \hat{w})=-\nabla_{x}\,\|x-w\|^2_2 = 2(\hat{w}-\hat{x}).    
\end{equation*}
Using this formula, we can write the internal gradient for the perturbed prediction $\hat{w}+t\Delta\hat{w}$ at the same point $\hat{x}$:
\begin{equation*}
    \nabla_{x}f_{QP}(\hat{x}, \hat{w} + t\Delta\hat{w}) = \nabla_{x}f_{QP}(\hat{x}, \hat{w}) + 2t\Delta\hat{w}. 
\end{equation*}
By definition, $\mathcal{N}(\hat{x})$ is a linear span of the vectors $\{ n_i|i\in I(\hat{x})\}.$ Hence, since $\Delta\hat{w}\in\mathcal{N}(\hat{x}),$ it can be expressed as 
\begin{equation*}
\Delta\hat{w}=\sum_{i\in I(\hat{x})}\delta_in_i,\quad\delta_i\in\R.
\tag{$\ast$}
\end{equation*}
 
By Property 2, the internal gradient has the following representation: 
\begin{equation*}
    \nabla_{x}f_{QP}(\hat{x}, \hat{w}) = \sum_{i\in I(\hat{x})}\alpha_in_i,\quad \alpha_i>0.
    \tag{$\ast\ast$}
\end{equation*}
Then, combining $(\ast)$ and $(\ast\ast),$ we obtain
\begin{equation*}
 \nabla_{x}f_{QP}(\hat{x}, \hat{w} + t\Delta\hat{w}) = \nabla_{x}f_{QP}(\hat{x}, \hat{w}) + 2t\Delta\hat{w} =\\ \sum_{i\in I(\hat{x})}(\alpha_i + 2t\delta_i)n_i
\end{equation*}
Since $\alpha_i>0,\,\forall i \in I(\hat{x})$, there exists $\epsilon>0,$ such that $\alpha_i -2t\delta_i > 0$ for $|t|<\epsilon.$ Therefore, $\nabla_{x}f_{QP}(\hat{x}, \hat{w} + t\Delta\hat{w})$ lies in the gradient cone of $\hat{x},$ and hence, by Property 2, $x_{QP}^\ast(\hat{w}+t\Delta\hat{w})=\hat{x}$ for $|t|<\epsilon.$ Therefore, the directional derivative of $x_{QP}^\ast(\hat{w})$ along $\Delta\hat{w}\in\mathcal{N}(\hat{x})$ is zero.
\paragraph{2. $\Delta\hat{w}\perp\mathcal{N}(\hat{x}).$ } Next, let $\Delta\hat{w}$ be orthogonal to $\mathcal{N}(\hat{x}).$ We begin with the first order approximation of $\hat{x}'(t):$
\begin{equation*}
    \hat{x}'(t)=\hat{x} + td + o(t).
\end{equation*} 
From the proof of Lemma 3, we can know that $d\perp \mathcal{N}.$ By definition of $x^\ast_{QP}$, we know that 
$\hat{x}$ is the point on $\mathcal{C}$ closest to $\hat{w}.$ Likewise, $\hat{x}'(t)$ is the point on $\mathcal{C}$ closest to $\hat{w}+t\Delta\hat{w}.$ Hence, $d=\Delta\hat{w}.$ Therefore, for any $\Delta\hat{w}\perp\mathcal{N},$ the directional derivative of $x_{QP}(\hat{w})$ along $\Delta\hat{w}$ is one. 

So, we have shown that 
\begin{equation*}
    \nabla_{\hat{w}}\,x^\ast_{QP}(\hat{w})\,\Delta\hat{w}=\begin{cases}
0 &\text{for } \Delta\hat{w}\in\mathcal{N}(\hat{x}) \\
\Delta\hat{w} & \text{for }  \Delta\hat{w}\perp\mathcal{N}(\hat{x}).
\end{cases}
\end{equation*}
Therefore, the lemma is proven.
\end{proof} 

\begin{proof}[Proof of Theorem 9]
First, we want to construct an orthogonal basis $\{e_1,\ldots e_n\}$ of $\R^n$ that will greatly simplify the calculations. We start by including the internal gradient in this basis, i.e., we define $e_1=\nabla_{x}f_{QP}(\hat{x}, \hat{w}).$ Then, let $I(\hat{x})=\{i|g_i(\hat{x})=0\}$ be the set of indices of the active constraints of the original problem and let $\mathcal{N}(\hat{x})=span(\{n_i|i\in I(\hat{x})\})$ be a linear span of their normals. By the liner independence condition from Assumption 2, $dim\big(\mathcal{N}(\hat{x})\big)=|I(\hat{x})|.$ Moreover, by Property 2, we know that $e_1\in\mathcal{N}(\hat{x}).$ Then, we can choose vectors $e_2,\ldots, e_{|I(\hat{x})|}$ that complement $e_1$ to an orthogonal basis of $\mathcal{N}(\hat{x}).$
The remaining vectors $e_{|I(\hat{x})| +1},\ldots,e_n,$ are chosen to complement $e_1,\ldots,e_{|I(\hat{x})|}$ to an orthogonal basis of $\R^n$. The choice of this basis is motivated by Lemma 6: $e_1$ is a basis of the null-space of the $r-$smoothed Jacobian, $e_1,\ldots,e_{|I(\hat{x})|}$ form a basis of the null space of the true QP Jacobian, and the remaining vectors form a basis of space in which we can move $x^\ast_{QP}(\hat{w}).$

For brevity, let $f_x=\nabla_{x}f(\hat{x}, w)$ denote the true gradient vector. By definition,
${\Delta\hat{w}=f_x\,\nabla_{\hat{w}}x^\ast_r(\hat{x}, \hat{w})}$
is obtained via the $r-$smoothed problem.
From Property 8, we know that $\Delta\hat{w}$ is a projection of $f_x$ on the vectors $e_2,\ldots,e_n.$ Then, since $e_1,\ldots,e_n$ is an orthogonal basis, we have
\begin{equation*}
    \Delta\hat{w}=\sum_{i=2}^n\beta_ie_i,\quad \beta_i=f_x^\top e_i,\, i=2,\ldots,n.
\end{equation*}

Now, let's see how this $\Delta\hat{w}$ affects the true decision $x^\ast_{QP}(\hat{w} + t\Delta\hat{w})$ for $t\to 0^+.$ First, we have a first-order approximation 
\begin{equation*}
x^\ast_{QP}(\hat{w} + t\Delta\hat{w})=\hat{x} + td + o(t),
\end{equation*}
for some $d\in\R.$ From Lemma 6, we know that $d$ is actually a projection of $\Delta\hat{w}$ onto the vectors $e_{|I(\hat{x})|+1},\ldots,e_n.$ 
Therefore, we have
\begin{equation*}
    x^\ast_{QP}(\hat{w} + t\Delta\hat{w})=\hat{x} + \sum_{i=|I(\hat{x}|+1}^n\beta_ie_i + o(t).
\end{equation*}

Finally, the change in the true objective can be expressed as
\begin{equation*}
\begin{aligned}
f\Big(x^\ast_{QP}(\hat{w}+t\Delta\hat{w}), w\Big)-f\Big(x^\ast_{QP}(\hat{w}), w\Big)
& =
tf^\top_x \Big(\sum_{i=|I(\hat{x}|+1}^n\beta_ie_i\Big) + o(t)
= \\ &=
t\sum_{i=|I(\hat{x}|+1}^n\beta_if^\top_xe_i+o(t)
=
t\sum_{i=|I(\hat{x}|+1}^n\beta_i ^2 +o(t)\geq 0.
\end{aligned}
\end{equation*}

Therefore, perturbing prediction along $\Delta\hat{w}$ does not decrease the true objective $f(\hat{x}, w),$ and hence \begin{equation*}
    f\big(x^\ast_{QP}(\hat{w}+t\Delta\hat{w}), w\big)\geq f\big(x^\ast_{QP}(\hat{w}), w\big) 
\end{equation*}
for $t\to 0^+.$
\end{proof}

\section{Equality constraints}
Assumption 2 postulates that for any $x\in\mathcal{C},$ the gradients of active constraints, $\{\nabla_{x}g_i(x)|g_i(x)=0\},$ are linearly independent. Now, suppose we include equality constraints in our problem. e.g., we have a constraint $g^{eq}(x)\leq0$ and $-g^{eq}(x)\leq 0$ for some $g.$ Clearly, the gradients of $g^{eq}(x)$ and $-g^{eq}(x)$ violate the independence assumption. However, we claim that it does not affect our results.
Let $\hat{w}$ and $\hat{x}$ be a prediction and a corresponding decision and let $n^{eq}=\nabla_{x}\,g^{eq}(\hat{x}).$ Suppose the equality constraint $g^{eq}(\hat{x})=0$ is active. Let $I(\hat{x})$ be the set of indices of the active constraints \textit{not including} $g^{eq}(x).$ Then, we have a representation of the internal gradient, 
\begin{equation*}
    \nabla_{x}f(\hat{x},\hat{w})=\alpha^{eq}_1n^{eq} - \alpha^{eq}_2n^{eq} + \sum_{i\in I(\hat{x})}\alpha_in_i.
\end{equation*}
Suppose that $\alpha^{eq}_1\neq\alpha_2^{eq},$ e.g., without loss of generality, $\alpha^{eq}_1>\alpha^{eq}_2.$ Then, 
\begin{equation*}
    \nabla_{x}f(\hat{x},\hat{w})=(\alpha^{eq}_1-\alpha^{eq}_2)n^{eq} + \sum_{i\in I(\hat{x})}\alpha_in_i
\end{equation*} and hence removing the constraint $-g^{eq}(x)\leq 0$ would not change the optimality of $\hat{x}.$ The remaining problem would satisfy complementary slackness and hence would have all the properties demonstrated in Section 3. Therefore, for the case with equality constraints, we need to extend the complementary slackness conditions by demanding $\alpha_1^{eq}\neq\alpha_2^{eq}.$ 

\section{Experimental details}
In this section, we provide the details of the experiments reported in the paper. All experiments were conducted on a machine with 32gb RAM and NVIDIA GeForce RTX 3070. The code is written in Python 3.8, and neural networks are implemented in \textit{PyTorch} 1.11. 
For methods requiring differentiation of optimization problems, we use the implementation by Agrawal et.\ al [2019]. The linear methods (SPO+, perturbed optimizers, identity-with-projection) were re-implemented by us. 

Experimental results reported in Figures 3 and 4 show the average and the standard deviation (shaded region) of the measured quantities across 4 random seeds.
For each seed, we randomly split data into train, validation, and test sets by using 70\%, 20\%, and 10\% of the whole dataset respectively. In Figure 3a, for each method and at each run, we take the model version corresponding to the best performance on the validation set and report its performance on the test set. In Figure 4, we do the same procedure at each training iteration.

In all experiments and methods, the predictor $\phi_\theta$ is represented by a fully connected neural network with two hidden layers of 256 neurons each, and \textit{LeakyReLU} activation functions. The output layer has no activation function. Instead, the output of the neural network is scaled by the factor $x_{scale}$ and shifted by $x_{shift}$. For linear methods and methods using QP approximation, the output layer predicts vector $\hat{w}$ of the same dimensionality as the decision variable. For the method using the true model, the prediction size is defined by the number of unknown parameters in the true objective function.
For training, we used the \textit{Adam} optimizer from PyTorch, with custom learning rate and otherwise default parameters.

Hyperparameters of all methods were chosen based on the results of the grid search reported in Tables 2-6.
Configuration files to reproduce the experiments and the code can be found at \textit{placeholder for GitHub link. For the reviewers, the code is submitted as an archive.}
\subsection{Portfolio optimization problem}
\begin{table}[t]
\centering
\begin{tabular}{p{3cm}|p{3cm}|p{1.5cm}|}
    {Parameter} & Search space & Best value \\
    \hline
    Learning rate & $\{0.5, 1, 5, 10\}\times 10^{-5}$ & $5\times 10^{-5}$\\
    Training epochs & $\{40, 80, 160\}$ & $80$\\
    Batch size & $\{1, 4, 8 ,32\}$ &  $1$\\
    $x_{shift}$ & $\{0, 0.1, 1\}$ & $0.1$ \\
    $x_{scale}$ & $\{0.1, 1\}$ & $1$\\
\end{tabular}
\caption{Hyperparameters for methods from Figure 3 for standard portfolio optimization problem with different $\lambda$'s.}
\end{table}

\begin{table}[t]
\centering
\begin{tabular}{p{3cm}|p{3cm}|p{1.5cm}|}
    {Parameter} & Search space & Best value \\
    \hline
    Learning rate & $\{0.5, 1, 5, 10\}\times 10^{-5}$ & $5\times 10^{-5}$\\
    Batch size & $\{1, 4, 8 ,32\}$ &  $1$\\
    $x_{shift}$ & $\{0, 0.1, 1\}$ & $0.1$ \\
    $x_{scale}$ & $\{0.1, 1\}$ & $0.1$\\
\end{tabular}
\caption{Hyperparameters for methods from Table 1 for LogSumExp portfolio optimization problem.}
\end{table}

\begin{table}[t]
\centering
\begin{tabular}{p{3cm}|p{3cm}|p{1.5cm}|}
    {Parameter} & Search space & Best value \\
    \hline
    Learning rate & $\{0.5, 1, 5, 10\}\times 10^{-5}$ & $5\times 10^{-5}$\\
    Batch size & $\{1, 4, 8 ,32\}$ &  $1$\\
    $x_{shift}$ & $\{0, 0.1, 1\}$ & $0.1$ \\
    $x_{scale}$ & $\{0.1, 1\}$ & $0.1$\\
    ID: projection & \{mean, norm, both\} & norm \\ 
    Perturbed: n samples & $\{1, 4, 8, 32\}$ & $4$ \\
    Perturbed: $\sigma$ & $\{0.05, 0.1, 0.5\}$ & $0.05$ \\
\end{tabular}
\caption{Hyperparameters for linear methods from Figure 4a, b for standard portfolio optimization problem with $\lambda=0$ and $\lambda=0.1$.}
\end{table}

Following Wang et.\ al [2020], we use historical data from QUANDL WIKI prices \cite{quandl} for 505 largest companies on the American market for the period 2014-2017. The dataset is processed and for every day we obtain a feature vector summarizing the recent price dynamic. For further details on the processing, we refer readers to the code and to  Wang et.\ al [2020]. 
The processed dataset contained historical data describing the past price dynamics for each of the 503 securities. For every random seed, 50 securities (thus, 50 decision variables) were chosen randomly.
The experiments on the LogSumExp variation of the portfolio optimization problem were conducted similarly. The hyperparameters for normal and LogSumExp portfolio problems are reported in Tables 2, 3, and 4.

\subsection{Optimal power flow problem}
\begin{figure}[h]
    \centering
    \includegraphics[scale=0.53]{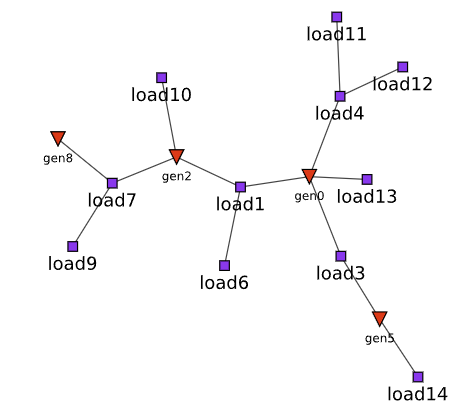}
    \caption{Example of randomly generated grid topology. Red triangles represent generator nodes, and purple squares represent loads.}
    \label{fig:grid}
\end{figure}
We considered a linearized DC-OPF problem that represents a DC grid without power losses. The decision variable is the vector of nodal voltages $v\in\R^n,$ and the unknown parameter $w$ represents either the value gained by serving power to a customer or the price paid for utilizing a generator. The reference voltage $v_0\in\R,$ the admittance matrix $Y,$ and the constraint bounds represent the physical properties of the grid. 
\begin{equation}
    \begin{aligned}
    \max_{v} &\quad& f(v, w)= -v_0w^\top (Yv)&& & \notag \\
    \text{subject to:} &\quad& \ushort{V}\leq v \leq \bar{V} && & \\ 
    &\quad&  \ushort{P} \leq-v_0Yv \leq \bar{P} && & \\
    &\quad&  \ushort{I} \leq Y_{ij}(v_i-v_j) \leq \bar{I} && & \\
    \end{aligned}
\end{equation}
Data for the DC OPF problem is generated artificially. First, we randomly generate a grid topology, see Figure~\ref{fig:grid} for an example. For each line, its admittance is set to $6S$. Nodal voltages are bounded between 325V and 375V, and the reference node has a fixed voltage of $v_0=350V.$ The demand in loads (power upper-bound), generators capacity (power lower-bound), and line current limits are sampled randomly from the following normal distributions: $\mathcal{N}(8000, 2500)\times$watt-hour, $\mathcal{N}(-14000, 2500)\times$watt-hour, $\mathcal{N}(25, 5)\times$ampere.
The coefficients $w$ are also sampled form the normal distributions: $\mathcal{N}(1.2, 1)$ for loads, and $\mathcal{N}(0.8, 0.1)$ for generators. Finally, all values are normalized such that $v_0$ becomes $7V$ (surprisingly, it performed better numerically than scaling $v_0$ to $1V$).
The observations $o$ is a concatenation of the true coefficients $w,$ demand of the loads, the capacity of the generators, and line current limits plus normally distributed noise with mean 0 and standard deviation 0.5.

\begin{table}[t]
\centering
\begin{tabular}{p{3cm}|p{3cm}|p{1.5cm}|}
    {Parameter} & Search space & Best value \\
    \hline
    Learning rate & $\{0.5, 1, 5, 10\}\times 10^{-5}$ & $5\times 10^{-5}$\\
    Batch size & $\{1, 4, 8 ,32\}$ &  $1$\\
    $x_{shift}$ & $\{0, 0.1, 1\}$ & $0.1$ \\
    $x_{scale}$ & $\{0.1, 1\}$ & $0.1$\\
    ID: projection & \{mean, norm, both\} & norm \\ 
    Perturbed: n samples & $\{1, 4, 8, 32\}$ & $4$ \\
    Perturbed: $\sigma$ & $\{0.05, 0.1, 0.5\}$ & $0.05$ \\
\end{tabular}
\caption{Hyperparameters for linear methods from Figure 4c for power flow problem.}
\end{table}

\subsection{Knapsack problem}
For the knapsack problem, we used the data generation process from the PyEPO package (Tang and Khalil 2022). The true problem is defined by the number of decision variables $n,$ objective coefficients $w\in\mathbb{R}^n,$ and $m$ resource constraints represented by $W\in\mathbb{R}^{n\times n}, b\in\R^m:$

\begin{equation}
    \begin{aligned}
    \max_{x} &\quad& f(x, w)= w^\top x&& & \notag \\
    \text{subject to:} &\quad& 0 \leq x \leq 1 && & \\ 
    &\quad&  Wx \leq b && &
    \end{aligned}
\end{equation}
In the experiments, we used $n=20, m=15$. Weights $W$ were sampled uniformly from the interval $(3, 8)$ and $b$ was sampled uniformly from $(20, 80).$ The constraints are fixed for all instances of the dataset (i.e., defined by the random seed).
For each seed, we construct a dataset of 512 samples. To do so we first sample the observation, $o\sim \mathcal{N}(0, 1),\: o\in\mathbb{R}^{512\times 16}$. Then, for each sample $i=\{1, \ldots, 512\}$, we compute the coefficients $w$ as
$$w^i=\big((\frac{1}{4} Bo^i + 3)^2 + 1\big)*0.4 + \epsilon^i,$$
where $B\in\R^{n\times p}$ is sampled from the Bernoulli distribution with $k=1, p=0.5$ (fixed across samples) and $\epsilon\in\mathbb{R}^{512\times 20}$ is noise uniformly sampled from $(0.5, 1.5)$ (different for each sample).
\begin{table}[t]
\centering
\begin{tabular}{p{3cm}|p{3cm}|p{1.5cm}|}
    {Parameter} & Search space & Best value \\
    \hline
    Learning rate & $\{0.5, 1, 5, 10\}\times 10^{-5}$ & $5\times 10^{-5}$\\
    Batch size & $\{1, 4, 8 ,32\}$ &  $1$\\
    $x_{shift}$ & $\{0, 0.1, 1\}$ & $0.1$ \\
    $x_{scale}$ & $\{0.1, 1\}$ & $0.1$\\
    ID: projection & \{mean, norm, both\} & norm \\ 
    Perturbed: n samples & $\{1, 4, 8, 32\}$ & $4$ \\
    Perturbed: $\sigma$ & $\{0.05, 0.1, 0.5\}$ & $0.05$ \\
\end{tabular}
\caption{Hyperparameters for linear methods from Figure 4d for the knapsack problem}
\end{table}

\end{document}